\documentclass{article} 
\usepackage{salesforce,times}

\usepackage[utf8]{inputenc}
\usepackage[T1]{fontenc}
\usepackage{amsmath}
\usepackage{nicefrac}
\usepackage{microtype}
\usepackage{arydshln}
\usepackage{adjustbox}

\usepackage{hyperref}
\usepackage{url}
\usepackage{bbm}
\usepackage{enumitem}
\usepackage{amsthm}
\usepackage{amssymb}
\usepackage{graphicx}
\usepackage{subcaption}
\usepackage{wrapfig}
\usepackage{algorithm}
\usepackage{algorithmic}
\usepackage{multirow}
\usepackage{booktabs}
\usepackage[table]{xcolor}
\usepackage{colortbl}
\usepackage{placeins}
\usepackage{titletoc}
\usepackage{cleveref}

\usepackage{amsmath,amsfonts,bm}

\def\eqref#1{equation~\ref{#1}}

\def\1{\bm{1}}

\DeclareMathAlphabet{\mathsfit}{\encodingdefault}{\sfdefault}{m}{sl}
\SetMathAlphabet{\mathsfit}{bold}{\encodingdefault}{\sfdefault}{bx}{n}

\newif\ifarxiv

\usepackage{xcolor}
\usepackage[breakable,skins,most]{tcolorbox}

\newtcbox{\oplabel}[2][]{
  on line,
  boxsep=1pt,
  left=3pt,
  right=3pt,
  top=1pt,
  bottom=1pt,
  arc=1pt,
  boxrule=0pt,
  colback=#2!12,
  coltext=#2!85!black,
  fontupper=\bfseries\scriptsize,
  #1
}
\definecolor{blockrow}{HTML}{EEF2F7}
\definecolor{insertpink}{HTML}{D9308A}
\definecolor{updategreen}{HTML}{188038}
\definecolor{deleteorange}{HTML}{F29900}

\definecolor{lightorange}{RGB}{245, 237, 211}
\definecolor{clovergreen}{RGB}{32,115,55}
\definecolor{redlinkcolor}{rgb}{0.79607843, 0.25098039, 0.25882353}
\definecolor{bluecitecolor}{rgb}{0,0.36,0.69}
\definecolor{lightorange}{RGB}{245, 237, 211} 
\definecolor{bluebar}{RGB}{138,159,201}
\definecolor{pinkbar}{RGB}{232,180,189}
\definecolor{deepgreen}{RGB}{152,206,187}
\definecolor{lightgreen}{RGB}{224,238,190}
\definecolor{googleblue}{RGB}{87,134,236}
\definecolor{stdgray}{gray}{0.5}
\newcommand{\valstd}[2]{$#1_{\,\textcolor{stdgray}{\scriptscriptstyle #2}}$}

\newcommand{\frozenmark}{%
  \raisebox{-0.12em}{\includegraphics[height=0.9em]{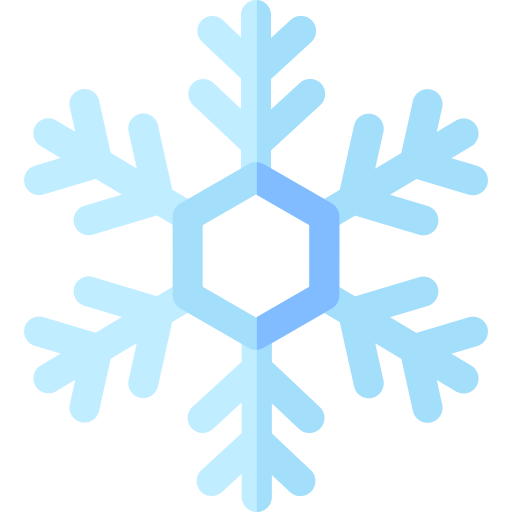}}%
}
\newcommand{\firemark}{%
  \raisebox{-0.12em}{\includegraphics[height=0.9em]{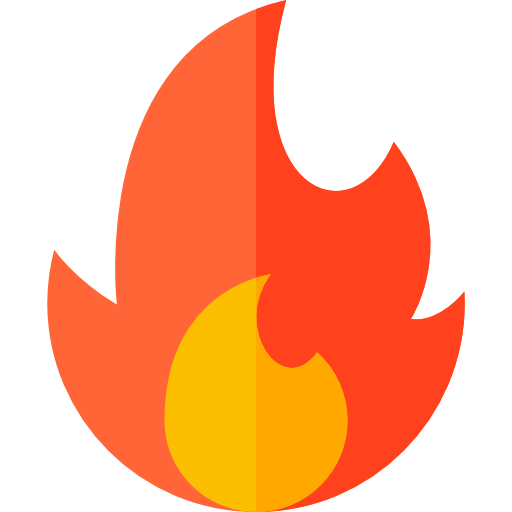}}%
}
\newcommand{\ourmethod}{\textsc{JitMem}\xspace} 
\newcommand{\skillos}{SkillOS\xspace}
\usepackage{xspace}

\newif\ifshowcomments
\showcommentstrue 
\ifshowcomments
\newcommand {\yefan}[1]{{\color{purple}\sf{[Yefan: #1]}}}
\newcommand {\yang}[1]{{\color{blue}\sf{[Yang: #1]}}}
\newcommand {\semih}[1]{{\color{orange}\sf{[Semih: #1]}}}
\newcommand {\shafiq}[1]{{\color{cyan}\sf{[Shafiq: #1]}}}

\newcommand {\addressedyang}[1]{{\color{blue}\sf{[(Addressed) Yang: #1]}}}
\newcommand {\addressedsemih}[1]{{\color{orange}\sf{[(Addressed) Semih: #1]}}}
\newcommand {\addressedshafiq}[1]{{\color{cyan}\sf{[(Addressed) Shafiq: #1]}}}
\else
\newcommand {\yefan}[1]{}
\newcommand {\yang}[1]{}
\newcommand {\semih}[1]{}
\newcommand{\shafiq}[1]{}
\newcommand {\addressedyang}[1]{}
\newcommand {\addressedsemih}[1]{}
\newcommand {\addressedshafiq}[1]{}
\fi

\usepackage{colortbl}
\usepackage{wrapfig}
\usepackage[table]{xcolor}

\usepackage{listings}
\usepackage{xcolor}
\usepackage{etoolbox} 
\newtcolorbox{prompt}{
  colback=black!2,    
  colframe=black!15,  
  boxrule=0.4pt,
  sharp corners,
  left=8pt,right=8pt,top=6pt,bottom=6pt,
  before skip=6pt, after skip=6pt,
  breakable,                 
  before upper=\ttfamily\small 
}

\definecolor{RubricNavy}{RGB}{16,36,132}    
\definecolor{RubricBack}{RGB}{235,238,249}  
\definecolor{RubricFrame}{RGB}{16,36,132}   

\newtcolorbox{rubricbox}[1]{%
  enhanced, breakable,
  colback=RubricBack,
  colframe=RubricFrame,
  colbacktitle=RubricNavy, coltitle=white,
  title=\sffamily\bfseries #1,
  boxrule=0.8pt, arc=2.5mm,
  left=8pt,right=8pt,top=8pt,bottom=8pt,
  before skip=8pt, after skip=8pt,
  before upper=\ttfamily\small 
                 \setlength{\parindent}{0pt}%
               \setlength{\parskip}{4pt}%

}

\setlist[itemize]{nosep, leftmargin=1.2em}
\setlist[enumerate]{nosep, leftmargin=1.5em}

\definecolor{hlred}{RGB}{200,0,0}

\newlist{trace}{description}{1}
\setlist[trace]{
    leftmargin=1.25cm,
    labelwidth=1.0cm,
    labelsep=0.25cm,
    font=\normalfont\bfseries,
    itemsep=2.5pt,
    topsep=3pt,
    parsep=0pt
}

\newtcolorbox{rhbox}[1]{
    colback=green!6,
    colframe=green!45!black,
    boxrule=0.4pt,
    fonttitle=\bfseries,
    title={#1},
    left=4pt,
    right=4pt,
    top=3pt,
    bottom=3pt,
    breakable,
    before skip=8pt,
    after skip=8pt
}

\arxivtrue

\ifarxiv
  \def\promptleft{-0.1cm}
  \def\promptright{-0.1cm}
\else
  \def\promptleft{-1cm}
  \def\promptright{-1cm}
\fi

\title{Just-in-Time Memory: Learning to Curate Task-Adaptive Memory for LLM Agents}

\sfsetauthors{Yefan Zhou*, Yang Li*, Zeyu Leo Liu, Semih Yavuz, Shafiq Joty}
\sfsetaffiliation{Salesforce AI Research}
\sfsetauthornote{\texttt{\{yefan.zhou,yli2,sjoty\}@salesforce.com}\\$^{*}$Equal contribution.\quad}

\begin{document}

\maketitle

\begin{abstract}

Agentic memory systems reuse past experience to improve future performance, yet most existing designs curate memory at write time: once a task is completed, its trajectory is distilled into a fixed artifact, such as a reflection, workflow, skill, or reasoning strategy, that is later retrieved by similarity. This forces the system to decide what is worth remembering before the future query is known, irreversibly discarding information and producing a query-independent summary that must serve many possible downstream tasks. Learning such a write-time curator is also difficult because the value of a storage decision may only become apparent when a relevant query arrives, potentially many tasks later, creating a long-horizon credit-assignment problem. We instead retain raw trajectories and defer curation until read time, when the current task is known. Given the retrieved traces and the new task, a memory curator synthesizes a compact, task-adaptive payload tailored to the immediate need. Because this payload is consumed on the same task, the curator can be trained directly from immediate task success, avoiding delayed utility signals and the need to artificially group related tasks. Across ALFWorld, WebShop, and $\tau^2$-bench, our Just-in-Time Memory (\ourmethod) consistently outperforms no-memory agents as well as heuristic and learned write-time memory methods, improving over the strongest baseline by 16.2, 16.3, and 3.9 absolute success-rate points, respectively. Notably, even an untrained curator is already competitive with or surpasses these baselines, showing that task-adaptive read-time curation itself is a major source of the gain; training the curator further compounds the improvement.\looseness-1

\end{abstract}
\ifarxiv\else\vspace{-5pt}\fi
\section{Introduction}
\label{sec:intro}
\ifarxiv\else\vspace{-3pt}\fi

Large language model (LLM) agents are increasingly expected to solve sequences of tasks that unfold over time, rather than isolated problems~\citep{wang2024survey,luo2026storage}. Starting from scratch on each task wastes one of the agent’s most valuable resources: its own prior experience. This has motivated a broad literature on agentic memory, which persists information from past trajectories and reuses it to improve future behavior~\citep{shinn2023reflexion,zhao2024expel,wang2023voyager,wang2024agent,ouyang2026reasoningbank}. Despite substantial variation in design, these methods share a common objective: memory is useful only insofar as it improves future task performance. Taking this future-utility perspective, we ask a more fundamental question: \emph{when} in the agent lifecycle should memory be shaped to best serve that objective?

\begin{figure*}[!th]
\centering
    \includegraphics[width=\linewidth]{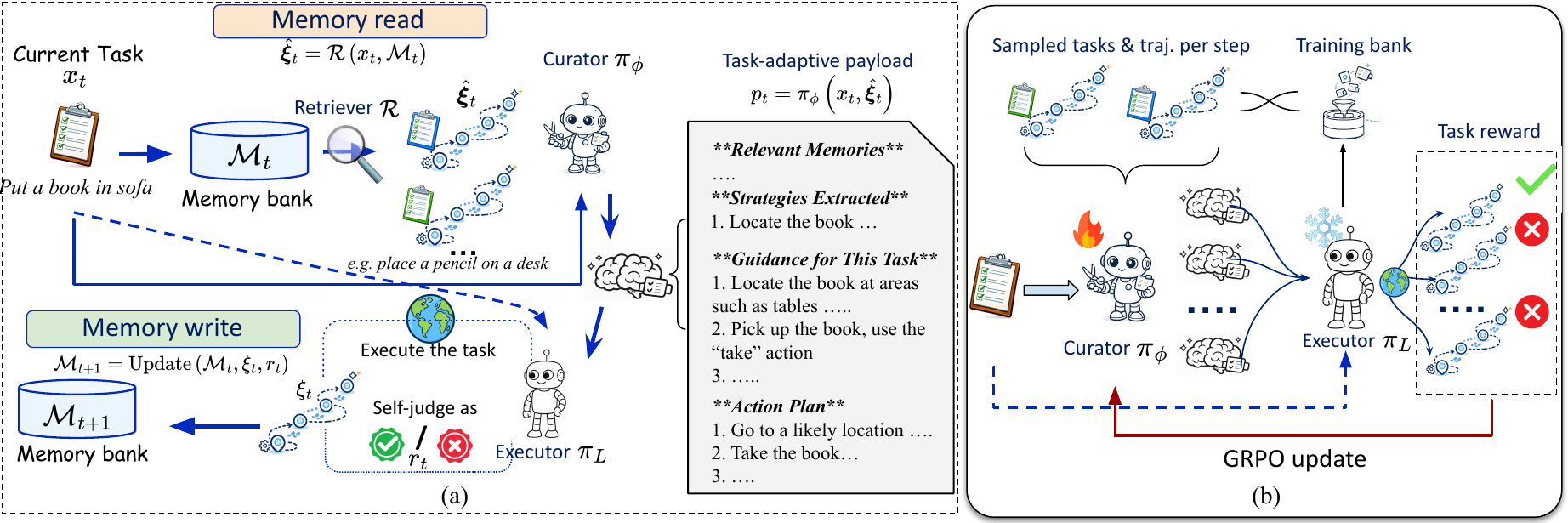}
    \ifarxiv\else\vspace{-16pt}\fi
    \caption{Overview of \ourmethod. (a) \textbf{Inference pipeline.} Given the current task $x_t$, the retriever fetches raw trajectories from the memory bank. The curator distills them conditioned on $x_t$ into a task-adaptive payload, which is injected into the executor's context. After execution, an executor-as-judge assesses correctness, and successful trajectories are stored back to the bank. (b) \textbf{Training pipeline.} At each training step, a task is sampled and relevant trajectories are retrieved from a fixed training bank, then the curator generates multiple candidate payloads. The frozen executor attempts the task with each payload and returns the immediate task reward, which is used to update the curator via GRPO.\looseness-1 
    }
    \label{fig:overview}
    \vspace{-11pt}
\end{figure*}

The dominant approach is to curate memory at write time. Once a task is completed, the system inspects the resulting trajectory and distills it into a persistent artifact such as verbal reflections~\citep{shinn2023reflexion}, natural-language insights~\citep{zhao2024expel}, reusable workflows~\citep{wang2024agent}, executable skills~\citep{wang2023voyager,ouyang2026skillos}, or transferable reasoning strategies~\citep{ouyang2026reasoningbank,fang2026memp}. At inference time, the agent retrieves one or more such artifacts, typically via similarity search, and incorporates them into its context. Crucially, however, the memory artifact is already fixed before the future query is known.

Curating memory at write time forces the system to decide what matters before the future task is known. This creates two fundamental costs. First, information loss is premature and irreversible: once details are discarded, a later task that depends on them has no way to recover them. Second, a single fixed artifact must serve many different future queries, even though the same trajectory may be useful in different ways depending on the task. A household interaction, for example, might teach one task a state-transition pattern (e.g., heating or cooling an object), while providing another with an object-placement strategy. A trajectory thus may not contain a single lesson but many possible lessons, and which lesson matters depends on the downstream task, which is unknown at write time.

Both costs stem from the same root cause: curation happens before the downstream task is known. We instead defer curation until read time, just in time, when the task to be solved is known. The memory bank remains a passive episodic store of raw trajectories, with no information discarded at write time. When a new task arrives, a retriever selects relevant traces, and a memory curator jointly reads those traces and the current task to synthesize a compact, task-conditioned payload. Because the curator sees the task, it can extract exactly the information that is useful for that task; the same stored trajectory can therefore yield different payloads for different downstream queries. This design parallels the cognitive-science view that episodic memory is reconstructive rather than replayed, with retrieval shaped by current goals and cues~\citep{Schacter2007TheCN}. 

Read-time curation also simplifies learning. Since the curated payload is consumed immediately by the current task, the curator can be optimized directly against same-task success, reducing the credit-assignment problem to a single interaction rather than waiting for uncertain future utility. This avoids the need to group related tasks to manufacture a learning signal, as required by learned write-time curators such as~\citet{ouyang2026skillos}, whose ablations identify grouping as a major contributor to performance. \ourmethod instantiates this principle as an RL-trained read-time curator operating over a persistent streaming memory bank. Figure~\ref{fig:overview} provides an overview of the system.

We evaluate \ourmethod on ALFWorld, WebShop, and $\tau^2$-bench, where it outperforms all baselines by \textbf{16.2, 16.3, and 3.9} absolute success-rate (SR) points, respectively. Even without training, read-time curation is already competitive with or substantially better than write-time curation built on the same underlying model: on WebShop, for example, untrained \ourmethod-gemini reaches $61.0$ SR versus $41.0$ for \skillos when both use Gemini-2.5-Pro as curator and executor. This indicates that task-adaptive read-time curation is itself a major source of the gains. RL training then compounds the gains. The trained curator also transfers to stronger executors without retraining. Beyond accuracy, its compact payload reduces input tokens by $50.3\%$--$56.3\%$ and executor steps by $28.4\%$--$31.4\%$ relative to write-time methods. Ablations further show that task-conditioned curation, quality-filtered storage, and retention of raw trajectories each contribute independently to the overall performance.


Our contributions are as follows:
\begin{itemize}[leftmargin=*,itemsep=0pt,topsep=0pt]
\item \textbf{Read-time curation enables task-adaptive memory.} By deferring curation to read time, the curator sees the current task and can tailor its distillation accordingly. The same stored trajectory yields different payloads for different tasks, a property that write-time curators cannot provide.
\item \textbf{Read-time curation simplifies credit assignment.} Since the curated payload is consumed on the same task it was produced for, the curator's reward is immediate. This collapses credit assignment to a single step, eliminating the task-grouping scaffolds required by learned write-time curators.
\item \textbf{\ourmethod: a read-time memory curator.} We introduce \ourmethod, which stores raw trajectories losslessly and synthesizes task-conditioned payloads at read time via a curator trained with GRPO over a persistent streaming memory bank.
\item \textbf{Empirical validation and analysis.} Across ALFWorld, WebShop, and $\tau^2$-bench, \ourmethod outperforms all baselines, including RL-trained write-time curators. Our analysis suggests that task-adaptive curation at read time is the key driver of improvement.
\end{itemize}

\ifarxiv\else\vspace{-10pt}\fi
\section{Related Work}
\label{sec:related}
\ifarxiv\else\vspace{-10pt}\fi

\noindent \textbf{Heuristic write-time memory.}
The dominant approach in agentic memory stores a distilled artifact at the end of each task and retrieves it by similarity at inference. Systems differ in what they distill: verbal reflections~\citep{shinn2023reflexion}, extracted insights~\citep{zhao2024expel}, memory streams with periodic summarization~\citep{park2023generative}, executable skills~\citep{wang2023voyager}, induced workflows~\citep{wang2024agent}, memory items at multiple granularities~\citep{fang2026memp}, self-organizing linked notes~\citep{xu2026mem}, and reasoning strategies from both successes and failures~\citep{ouyang2026reasoningbank}. \citet{ma2026deserves} use prediction-error signals to decide which experiences deserve distillation, adding adaptivity to what is stored. Despite these differences, all share two properties: curation is triggered at write time, and the stored artifact is query-independent, fixed before any future task is seen. ReasoningBank~\citep{ouyang2026reasoningbank}, the most competitive recent instance, distills transferable reasoning strategies via a prompted LLM and retrieves them by cosine similarity.

\noindent \textbf{Learned write-time memory.}
A growing line trains the memory-writing policy directly. Retroformer~\citep{yao2024retroformer} fine-tunes a retrospective model to rewrite the agent's prompt, though it operates within a single task instance. Several recent methods train memory operations as RL-optimized actions: Memory-R1~\citep{yan2025memoryr1}, Agentic Memory~\citep{yu2026agentic} with a progressive GRPO curriculum, Memento~\citep{zhou2025memento} with a case-selection policy, and Memory as a Controlled Process~\citep{jiang2026memory} with a lightweight control policy. MemRefine~\citep{kim2026memrefine} compresses the stored bank offline via LLM-guided merging. All of these operate at write or maintenance time. SkillOS~\citep{ouyang2026skillos}, the closest prior work, trains a skill curator with GRPO, but must group related tasks to manufacture a delayed learning signal because the reward for a write decision arrives only when a future query matches. By moving curation to read time, \ourmethod makes the reward immediate and eliminates the need for task grouping.

\noindent \textbf{Learned in-session working memory.}
A related thread uses RL to manage the context window within a single task execution: Sculptor~\citep{li2026sculptor} and ContextCurator~\citep{li2026escaping} train policies to compress or restructure the accumulating observation history, MemSearcher~\citep{yuan2025memsearcher} iteratively rewrites a fixed-length working memory, and Proactive Memory Agent~\citep{wu2026remember} learns when to inject reminders during long-horizon tasks. Recuris~\citep{yu2026recursive} combines step-level working-memory selection with cross-task skill evolution. All optimize in-session or turn-level context; \ourmethod instead curates persistent episodic memory across tasks.

\noindent \textbf{Read-time and test-time context processing.}
A separate line constructs better context from accumulated experience at test time. Synapse~\citep{zheng2024synapse} retrieves full trajectories as exemplars but does not distill or condition on the incoming task. MemToolAgent~\citep{er2026memtoolagent} adapts how many entries to retrieve based on the similarity distribution, but the entries themselves are distilled at write time and returned unchanged. Decocted experience~\citep{shen2026decocted} distills past trajectories into lessons, but each lesson is distilled query-independently and the policy is prompted rather than learned. Agentic Plan Caching~\citep{zhang2026agentic} extracts reusable plan templates, though the plan structure is fixed at extraction. SkillTTA~\citep{wang2026skills} synthesizes a task-conditioned skill at test time via meta prompt optimization, but from a preconstructed pool that does not grow during deployment. MemHarness~\citep{wu2026memharness}, concurrent with our work, also curates at read time: it trains a single policy with GRPO that both adapts retrieved experience and executes the task; because curation and execution are entangled in one model, the trained policy does not transfer across executors. \ourmethod decouples the curator from the executor, enabling cross-executor transfer, and operates over a persistent streaming bank. See \citet{luo2026storage} for a broader survey.

\ifarxiv\else\vspace{-5pt}\fi
\section{Method}
\label{sec:method}
\ifarxiv\else\vspace{-5pt}\fi

In a \emph{streaming task setting}, an agent receives a sequence of tasks $\{x_1, x_2, \ldots, x_T\}$ one at a time. At each step $t$, the agent interacts with an environment to solve $x_t$, producing a trajectory $\xi_t = (o_1, a_1, \ldots, o_n, a_n)$ of interleaved observations $o_i$ and actions $a_i$, and receives a task-success reward $r_t \in [0, 1]$. \ourmethod builds on four components (Figure~\ref{fig:overview}): a \emph{memory bank} $\mathcal{M}_t$ that stores raw trajectories from past tasks, a retriever $\mathcal{R}$, a memory curator $\pi_\phi$, and a frozen agent executor $\pi_L$. Only the curator is trainable. The objective is to maximize expected cumulative task success $\max_\phi \;\mathbb{E}\!\left[\sum_{t=1}^{T} r_t\right]$. At each task, the pipeline operates in four steps:
\begin{enumerate}[leftmargin=*,noitemsep,topsep=0pt,partopsep=0pt,parsep=0pt]
\item \textbf{Retrieve:} $\hat{\boldsymbol{\xi}}_t = \mathcal{R}(x_t, \mathcal{M}_t)$, fetch the top-$k$ raw trajectories from the memory bank.
\item \textbf{Curate:} $p_t = \pi_\phi(x_t, \hat{\boldsymbol{\xi}}_t)$, synthesize a task-adaptive payload.
\item \textbf{Execute:} $(\xi_t, r_t) = \pi_L(x_t, p_t)$, run the frozen executor with $p_t$ in context.
\item \textbf{Update:} $\mathcal{M}_{t+1} = \textsc{Update}(\mathcal{M}_t, \xi_t, r_t)$, append $\xi_t$ to the bank if the quality gate accepts it.
\end{enumerate}
The curated payload $p_t$ is ephemeral and is not stored. Instead, only the resulting trajectory $\xi_t$ is considered for insertion into the memory bank. We describe each component below and then the training procedure.

\textbf{Memory Bank~}
The memory bank $\mathcal{M}$ stores complete, unabstracted trajectories. Each entry is a raw trajectory $\xi = (x, o_1, a_1, \ldots, o_n, a_n)$ comprising the task description and the full interleaved observation--action sequence. No summarization, reflection, or skill abstraction is applied at storage time. Preserving raw traces is essential: it allows the curator to extract different information from the same trajectory for different tasks, an affordance lost when trajectories are distilled to fixed summaries at storage. Since task success labels are unavailable at deployment, we use the executor model as LLM-as-judge~\citep{ouyang2026reasoningbank} to gate which trajectories enter the bank: \textsc{Update} appends $\xi_t$ only if the judge deems the task successfully solved. The intent is to keep retrieved demonstrations as positive exemplars. Section~\ref{sec:analysis} ablates this choice against storing all trajectories and labeling each as success or failure when presented to the curator.

\textbf{Retrieval~}
The retriever $\mathcal{R}$ selects the top-$k$ trajectories from $\mathcal{M}$ most relevant to the current task $x_t$. We use BM25~\citep{robertson2009probabilistic} over task descriptions only (not trajectory content), keeping retrieval lightweight and decoupled from trajectory length. We choose BM25 for consistency with baselines, and the framework places no constraint on the retriever. The retrieved trajectories are concatenated in ranked order and passed to the curator. The retriever is not trained and operates identically at training and test time. The choice of $k$ is reported in Appendix~\ref{sec:appendix_hypers}.

\textbf{Memory Curator~}
The curator's input is a structured prompt containing the current task description $x_t$ followed by the $k$ retrieved raw trajectories $\hat{\boldsymbol{\xi}}_t$, delimited by lightweight separators. Its output is a compact natural-language \emph{memory payload} $p_t = \pi_\phi(x_t, \hat{\boldsymbol{\xi}}_t)$: a concise briefing that identifies the most relevant past experiences, extracts strategies that worked on similar tasks, and provides specific guidance for the current task (prompt in Appendix~\ref{sec:appendix_hypers}). Since $p_t$ depends on $x_t$, the same retrieved trajectory yields a different distillation for each task that retrieves it — the \emph{task-adaptive} property central to our approach.

\textbf{Agent Executor~}
The executor $\pi_L$ is a frozen pretrained LLM that is never updated during curator training. Freezing the executor keeps the system modular: one trained curator can serve multiple executors without retraining, and the memory component can be evaluated in isolation. Given the current task $x_t$ and the curated payload $p_t$, the executor generates actions to solve the task. The payload is prepended to the executor's prompt, providing task-relevant guidance extracted from past experience (prompt in Appendix~\ref{sec:appendix_hypers}); the executor therefore acts from the compact curated payload rather than directly consuming the raw retrieved trajectories. The same executor model also serves as the LLM-as-judge for the memory update policy.

\textbf{Curator Training~}
To train the curator, we use GRPO~\citep{shao2024grpo}: for each sampled training task $x_t$, the retriever fetches trajectories $\hat{\boldsymbol{\xi}}_t$ from the memory bank and the curator generates a group of $G$ candidate payloads $\{p_t^{(i)}\}_{i=1}^G$. The frozen executor attempts $x_t$ with each payload and returns the ground-truth task reward $r_t^{(i)} \in [0, 1]$, the benchmark's native evaluation metric (binary success on ALFWorld and $\tau^2$-bench, continuous score on WebShop). GRPO computes per-group advantages $\hat{A}_i = r_t^{(i)} - \text{mean}_j\, r_t^{(j)}$ (we omit the standard-deviation normalization following \citet{liu2024understanding}) and updates $\pi_\phi$ via:
\ifarxiv\else\vspace{-3mm}\fi
\begin{equation*}\textstyle
  \mathcal{L}_{\text{GRPO}} = -\frac{1}{G}\sum_{i=1}^{G} \hat{A}_i \cdot \log \pi_\phi(p_t^{(i)} \mid x_t, \hat{\boldsymbol{\xi}}_t),
\end{equation*}
without a value network. The executor $\pi_L$ remains frozen throughout. The key property of this design is that $r_t$ is a direct function of the payload $p_t$ produced for that same task $t$, with no intervening steps: the temporal gap between the curator's action and its reward is zero. This makes credit assignment immediate and eliminates the need for task-grouping or delayed-return machinery. In write-time memory, by contrast, a storage decision at step $s$ is graded only when a future task $t > s$ retrieves the artifact, possibly many tasks later.

At deployment, the memory bank grows online as tasks are solved. For training, we want the curator's reward to reflect payload quality alone, not the stochasticity of which trajectories happen to be available. We therefore construct a fixed training bank by running the base executor (without the curator) on the training set once and retaining successful trajectories using ground-truth success labels rather than the LLM judge. This bank is held fixed throughout training, ensuring stable and reproducible learning. A mild train/test distribution shift results: the training bank contains base-executor trajectories, while at test time the bank grows with curator-augmented ones. Section~\ref{sec:analysis} studies a staged bank refresh to quantify and close this gap.

\textbf{Evaluation Procedure~}
By default, the memory bank is initialized \emph{empty} at the start of each test sequence; the training bank does not carry over. The bank grows organically as tasks are solved, so early tasks benefit less from memory than later ones, resulting in a natural cold-start effect. Section~\ref{sec:analysis} studies warm-starting the test bank with training-time trajectories to mitigate this. For evaluation efficiency, we use a batched streaming protocol: tasks within a batch share the same memory bank state, and the bank is updated after each batch. Task success is measured by the benchmark's ground-truth verifier, while the memory update policy uses the LLM judge to avoid leaking ground-truth labels into the bank. Since both task ordering and batch composition affect performance, we report results averaged over multiple runs with different random orderings (Section~\ref{sec:experiments}).

\ifarxiv\else\vspace{-5pt}\fi
\section{Experiments}
\label{sec:experiments}
\ifarxiv\else\vspace{-5pt}\fi

\begin{table*}[t]
\centering
\setlength{\tabcolsep}{5.5pt}
\small
\setlength{\belowcaptionskip}{6.0pt}
\caption{Results on ALFWorld and WebShop across three frozen executors. Gains in red are relative to the strongest baseline in each block. \frozenmark~denotes a prompted (untrained) curator; \firemark~denotes an RL-trained curator. Mean and standard deviation over 3 runs with different task orderings.}
\vspace{-3mm}
\label{tab:merged_results}
\resizebox{0.8\textwidth}{!}{%
\begin{tabular}{lllll}
    \toprule
    \multirow{2}{*}{\textbf{Methods}}
    & \textbf{Curator}
    & \textbf{ALFWorld}
    & \multicolumn{2}{c}{\textbf{WebShop}} \\
    \cmidrule(lr){3-3} \cmidrule(lr){4-5}
    &
    & \textbf{SR} & \textbf{Score} & \textbf{SR} \\
    \midrule
    \multicolumn{5}{c}{\cellcolor{gray!15}\textit{Executor: Qwen3-8B}} \\
    No Memory & ---
    & \valstd{47.9}{1.2} & \valstd{33.3}{0.7} & \valstd{\phantom{0}9.8}{0.5} \\
    ReasoningBank & \adjustbox{valign=c}{\frozenmark} Qwen3-8B
    & \valstd{55.7}{3.1} & \valstd{35.4}{1.1} & \valstd{11.4}{0.9} \\
    MemP & \adjustbox{valign=c}{\frozenmark} Qwen3-8B
    & \valstd{49.7}{0.7} & \valstd{35.7}{0.9} & \valstd{12.0}{0.5} \\
    \skillos{}-base & \adjustbox{valign=c}{\frozenmark} Qwen3-8B
    & \valstd{53.1}{2.5} & \valstd{38.6}{0.9} & \valstd{13.6}{0.8} \\
    \skillos{}-gemini & \adjustbox{valign=c}{\frozenmark} Gemini-2.5-Pro
    & \valstd{50.7}{3.6} & \valstd{38.1}{1.0} & \valstd{13.2}{0.9} \\
    \skillos{} & \adjustbox{valign=c}{\firemark} Qwen3-8B
    & \valstd{61.2}{4.6} & \valstd{40.6}{0.7} & \valstd{16.5}{0.7} \\
    \hdashline
    \rowcolor{blockrow} \ourmethod{}-base  & \adjustbox{valign=c}{\frozenmark} Qwen3-8B
    & \valstd{60.5}{2.6} & \valstd{32.5}{3.2} & \valstd{11.7}{0.5} \\
    \rowcolor{blockrow} \ourmethod{} & \adjustbox{valign=c}{\firemark} Qwen3-8B
    & \valstd{77.4}{2.9}(\textcolor{red}{+16.2})  & \valstd{61.1}{0.9}(\textcolor{red}{+20.5}) & \valstd{32.8}{1.7}(\textcolor{red}{+16.3}) \\ 
    \midrule
    \multicolumn{5}{c}{\cellcolor{gray!15}\textit{Executor: Gemini-2.5-Pro}} \\
    No Memory & ---
    & \valstd{66.4}{2.0} & \valstd{48.6}{0.3} & \valstd{38.4}{0.5} \\
    ReasoningBank & \adjustbox{valign=c}{\frozenmark} Qwen3-8B
    & \valstd{71.4}{2.9} & \valstd{47.1}{1.0} & \valstd{38.0}{0.6} \\ 
    ReasoningBank & \adjustbox{valign=c}{\frozenmark} Gemini-2.5-Pro
    & \valstd{78.6}{2.9} & \valstd{50.8}{1.5} & \valstd{40.2}{1.3} \\
    MemP & \adjustbox{valign=c}{\frozenmark} Qwen3-8B
    & \valstd{74.3}{3.4} & \valstd{51.9}{1.9} & \valstd{40.3}{1.3} \\
    MemP & \adjustbox{valign=c}{\frozenmark} Gemini-2.5-Pro
    & \valstd{77.1}{2.1} & \valstd{51.3}{1.2} & \valstd{39.8}{1.0} \\
    \skillos{}-base & \adjustbox{valign=c}{\frozenmark} Qwen3-8B
    & \valstd{70.7}{3.0} & \valstd{52.8}{1.0} & \valstd{39.6}{0.8} \\
    \skillos{}-gemini & \adjustbox{valign=c}{\frozenmark} Gemini-2.5-Pro
    & \valstd{79.3}{2.6} & \valstd{54.7}{1.0} & \valstd{41.0}{1.2} \\
    \skillos{} & \adjustbox{valign=c}{\firemark} Qwen3-8B
    & \valstd{80.2}{3.1} & \valstd{56.0}{0.7} & \valstd{41.3}{0.8} \\
    \hdashline
    \rowcolor{blockrow} \ourmethod{}-base  & \adjustbox{valign=c}{\frozenmark} Qwen3-8B
    & \valstd{80.0}{1.5} & \valstd{54.7}{1.4} & \valstd{44.4}{0.6}  \\ 
    \rowcolor{blockrow} \ourmethod{} & \adjustbox{valign=c}{\firemark} Qwen3-8B
    & \valstd{86.2}{1.9}(\textcolor{red}{+6.0})  & \valstd{61.0}{0.8}(\textcolor{red}{+5.0}) & \valstd{50.5}{0.8}(\textcolor{red}{+9.2}) \\ 
    \rowcolor{blockrow} \ourmethod{}-gemini  & \adjustbox{valign=c}{\frozenmark} Gemini-2.5-Pro
    & \valstd{81.9}{2.7} & \valstd{72.1}{1.0} & \valstd{61.0}{0.7} \\
    \midrule
    \multicolumn{5}{c}{\cellcolor{gray!15}\textit{Executor: GPT-5.4}} \\
    No Memory & ---
    & \valstd{62.6}{0.3} & \valstd{40.9}{0.5} & \valstd{32.6}{0.6} \\
    ReasoningBank & \adjustbox{valign=c}{\frozenmark} Qwen3-8B
    & \valstd{69.8}{4.0} &  \valstd{37.4}{1.8} & \valstd{29.5}{1.0} \\
    ReasoningBank & \adjustbox{valign=c}{\frozenmark} GPT-5.4
    & \valstd{77.9}{4.2} &  \valstd{43.1}{1.7} & \valstd{33.6}{1.5} \\             
    MemP & \adjustbox{valign=c}{\frozenmark} GPT-5.4
    & \valstd{72.6}{1.7} & \valstd{40.8}{1.0} & \valstd{34.5}{1.7} \\
    \skillos-base & \adjustbox{valign=c}{\frozenmark} Qwen3-8B
    & \valstd{66.9}{1.8} & \valstd{39.7}{2.2} & \valstd{31.0}{2.0}  \\
    \skillos-gpt & \adjustbox{valign=c}{\frozenmark} GPT-5.4
    & \valstd{70.0}{3.3} & \valstd{33.3}{1.1} & \valstd{26.9}{1.4} \\
    \hdashline
    \rowcolor{blockrow} \ourmethod{}-base  & \adjustbox{valign=c}{\frozenmark} Qwen3-8B
    & \valstd{79.3}{3.6} &  \valstd{49.9}{1.0} & \valstd{39.3}{0.7} \\
    \rowcolor{blockrow} \ourmethod{}& \adjustbox{valign=c}{\firemark} Qwen3-8B
    & \valstd{86.7}{0.7}(\textcolor{red}{+8.8}) &  \valstd{53.8}{0.3}(\textcolor{red}{+10.7}) & \valstd{45.4}{0.0}(\textcolor{red}{+10.9}) \\
    \rowcolor{blockrow} \ourmethod{}-gpt & \adjustbox{valign=c}{\frozenmark} GPT-5.4
    & \valstd{83.3}{2.1} & \valstd{57.0}{1.5} & \valstd{47.5}{1.4} \\
    \bottomrule
\end{tabular}
}
\vspace{-4mm}
\end{table*}

We evaluate on three agentic benchmarks: ALFWorld~\citep{shridhar2020alfworld} (text-based embodied control, 140 test tasks), WebShop~\citep{yao2022webshop} (web-based product purchase, 500 test instances), and $\tau^2$-bench~\citep{barres2025tau} (conversational tool-use across airline, retail, and telecom domains). We report success rate (SR) on all three and additionally averaged score on WebShop.

We compare against a no-memory agent (the frozen executor $\pi_L$ alone) and three write-time memory baselines: ReasoningBank~\citep{ouyang2026reasoningbank}, which distills strategies and insights from past experiences; MemP~\citep{fang2026memp}, which generates memory items at multiple granularities; and \skillos~\citep{ouyang2026skillos}, which trains a skill curator via RL with composite rewards on grouped task streams. For both \skillos and \ourmethod, we include ``-base'' variants (same base model, without curator training) and ``-gpt/-gemini'' variants (prompted GPT-5.4 or Gemini-2.5-Pro as curator) to test whether a strong prompted model can serve as a zero-shot curator.

We evaluate with three frozen executors: Qwen3-8B, Gemini-2.5-Pro, and GPT-5.4. The trained curator is initialized from Qwen3-8B with thinking mode disabled and optimized with GRPO for 100 steps (learning rate $1\times10^{-6}$, batch size 32, group size 8), using Qwen3-8B as the executor during training for efficiency. The trained curator generalizes to stronger executors at test time (\Cref{tab:transfer}). The retriever is fixed across all methods and variants. At test time, tasks are processed in batches that share the same memory bank state, with the bank updated after each batch. We report mean $\pm$ standard deviation over multiple runs with different task orderings. See Appendix~\ref{sec:appendix_hypers} for complete setups.\looseness-1

\ifarxiv\else\vspace{-5pt}\fi
\subsection{Main Results}
\ifarxiv\else\vspace{-5pt}\fi

\textbf{Read-time curation outperforms write-time curation under the same zero-shot curator.}
To isolate curation strategy from curator capacity, we compare training-free variants of our method (\ourmethod-base, \ourmethod-gpt/gemini) against training-free write-time baselines using the same curator model. Across all three benchmarks (\Cref{tab:merged_results,tab:taubench_results}), the read-time variants generally outperform their write-time counterparts: for example, with Qwen3-8B as both executor and curator on ALFWorld, \ourmethod-base reaches 60.5 SR versus 55.7 for ReasoningBank and 53.1 for \skillos-base. The pattern holds with Gemini-2.5-Pro on WebShop (61.0 vs. 40.2/41.0) and GPT-5.4 on $\tau^2$-bench (75.6 vs. 71.7/66.4). Unlike ALFWorld and WebShop, $\tau^2$-bench requires multi-turn tool-use dialogues where the agent must converse with a user, call tools, and enforce domain policies. Per-domain gains are largest on Telecom (+11.0), where tasks involve complex multi-step policy verification. On Airline and Retail, none of the memory methods improves over the no-memory agent beyond variance, and \ourmethod variants remain on par with the baselines. This suggests that read-time curation is most valuable when tasks require synthesizing procedural guidance rather than simple fact retrieval. Our method with a weaker curator can even surpass baselines using a stronger one: with GPT-5.4 as executor on ALFWorld, \ourmethod-base with Qwen3-8B as curator (79.3) outperforms ReasoningBank (77.9) and \skillos-gpt (70.0), both using GPT-5.4. This confirms that gains stem from read-time task-adaptive curation rather than curator model strength.

\begin{table*}[t]
\centering
\setlength{\tabcolsep}{5.5pt}
\small
\setlength{\belowcaptionskip}{6.0pt}
\caption{Results on $\tau^2$-bench with GPT-5.4 as executor, reporting SR per domain and macro/micro averages. Mean and standard deviation over 4 runs with different task orderings.}
\label{tab:taubench_results}
\ifarxiv\else\vspace{-3mm}\fi
\resizebox{0.8\textwidth}{!}{%
\begin{tabular}{llcccll}
    \toprule
    \multirow{2}{*}{\textbf{Methods}}
    & \textbf{Curator}
    & \multicolumn{5}{c}{\textbf{$\tau^2$-bench}} \\
    \cmidrule(lr){3-7}
    & 
    & \textbf{Airline} & \textbf{Retail} & \textbf{Telecom} & \textbf{Macro Avg.} & \textbf{Micro Avg.} \\
    \multicolumn{7}{c}{\cellcolor{gray!15}\textit{Executor: GPT-5.4}} \\
    No Memory &  --- &  \valstd{65.0}{5.7}    &    \valstd{79.8}{4.2}         &     \valstd{50.2}{2.0}     &    \valstd{65.0}{3.2}   & \valstd{65.0}{2.8}    \\
    ReasoningBank &  \adjustbox{valign=c}{\frozenmark} Qwen3-8B  &  \valstd{67.0}{5.7}    &    \valstd{79.8}{3.1}         &     \valstd{46.5}{5.4}     &\valstd{64.4}{2.3}  &\valstd{63.8}{3.0}   \\
    ReasoningBank &  \adjustbox{valign=c}{\frozenmark} GPT-5.4  &  \valstd{65.0}{5.0}    &    \valstd{84.6}{2.2}         &     \valstd{61.6}{8.4}     &\valstd{70.4}{4.1}  &\valstd{71.7}{3.9}   \\
    \skillos{}-base &  \adjustbox{valign=c}{\frozenmark} Qwen3-8B  &  \valstd{68.5}{4.6}    &    \valstd{79.6}{2.7}         &     \valstd{52.0}{2.1}     &  \valstd{66.7}{1.8}  & \valstd{66.3}{1.3}  \\
    \skillos{}-gpt &  \adjustbox{valign=c}{\frozenmark} GPT-5.4  &  \valstd{68.0}{1.4}    &    \valstd{78.5}{3.5}         &     \valstd{53.3}{2.7}     &  \valstd{66.6}{1.3}  & \valstd{66.4}{1.9}  \\
    \hdashline 
    \rowcolor{blockrow} \ourmethod{}-base & \adjustbox{valign=c}{\frozenmark} Qwen3-8B
    & \valstd{66.0}{4.9} 
    & \valstd{81.1}{1.5} 
    & \valstd{57.9}{8.5} 
    & \valstd{68.3}{2.2} 
    & \valstd{68.9}{2.6} \\
    \rowcolor{blockrow} \ourmethod{}-gpt & \adjustbox{valign=c}{\frozenmark} GPT-5.4
    & \valstd{63.5}{3.0} 
    & \valstd{84.0}{3.6} 
    & \valstd{72.6}{4.5} 
    & \valstd{73.4}{2.0}(\textcolor{red}{+3.0}) 
    & \valstd{75.6}{2.3}(\textcolor{red}{+3.9}) \\
    \bottomrule
\end{tabular}
}
\ifarxiv\else\vspace{-6mm}\fi
\end{table*}

\textbf{Learned read-time curation surpasses learned write-time curation.}
When curators are RL-trained with the same Qwen3-8B base model (\Cref{tab:merged_results}, first block), \ourmethod outperforms \skillos by a large margin: 77.4 vs. 61.2 (+16.2) on ALFWorld and 32.8 vs. 16.5 (+16.3) SR on WebShop. The gap persists with stronger executors: with Gemini-2.5-Pro, \ourmethod reaches 86.2 vs. 80.2 (+6.0) on ALFWorld and 50.5 vs. 41.3 (+9.2) on WebShop. Notably, this improvement comes with a simpler training setup: we use only the task reward, while \skillos requires an additional judge model to assign content-quality rewards and groups related tasks to create temporal dependencies. We provide RL training curves (\Cref{fig:training-curve-alfworld,fig:training-curve-webshop}) in Appendix~\ref{sec:appendix_results}. We evaluate only training-free variants of \ourmethod and baselines on $\tau^2$-bench, since the benchmark does not provide a standard training split and designing effective synthetic training data for it remains an open problem.\looseness-1

\begin{wraptable}{r}{0.44\textwidth}
\centering
\scriptsize
\caption{Executor transfer on ALFWorld (SR). Columns are test-time executors.}
\label{tab:transfer}
\vspace{-5pt}
\begin{tabular}{lcc}
\toprule
\textbf{Curator training} & \textbf{Qwen3-8B} & \textbf{GPT-5.4} \\
\midrule
No training                    & \valstd{60.5}{2.6} & \valstd{79.3}{3.6} \\
Trained w/ Qwen3-8B exec.     & \valstd{77.4}{2.9} & \valstd{86.7}{0.7} \\
Trained w/ GPT-5.4 exec.      & ---                & \valstd{88.1}{0.9} \\
\midrule
Transfer gap                   & ---                & 1.4 \\
\bottomrule
\end{tabular}
\end{wraptable}
\textbf{The learned curator transfers across executors.}
The \ourmethod curator is trained once with Qwen3-8B as executor, yet transfers to stronger executors without retraining. As shown in \Cref{tab:transfer}, the transferred curator closes to within 1.4 SR points of one trained directly with GPT-5.4, 
suggesting it learns generalizable curation strategies rather than executor-specific patterns. Across both Gemini-2.5-Pro and GPT-5.4 executors (\Cref{tab:merged_results}), the transferred curator consistently improves over \ourmethod-base (+6.2/+7.4 on ALFWorld, +6.1 on WebShop) and outperforms RL-trained \skillos on Gemini-2.5-Pro. A single trained curator can thus serve multiple executors, reducing deployment costs.

\begin{wraptable}{r}{0.44\textwidth}
\centering
\scriptsize
\setlength{\tabcolsep}{3pt}
\ifarxiv\else\vspace{-4mm}\fi
\caption{Average tokens (K) and steps per task on ALFWorld with GPT-5.4 executor.}
\label{tab:efficiency}
\ifarxiv\else\vspace{-5pt}\fi
\begin{tabular}{lccc}
\toprule
\textbf{Method} & \textbf{In. Tok.} ($\downarrow$) & \textbf{Out. Tok.} ($\downarrow$) & \textbf{Steps} ($\downarrow$) \\
\midrule
No Memory     & \phantom{0}9.0 & 1.40 & 17.8 \\
ReasoningBank & 19.7 & 1.26 & 16.2 \\
\skillos-base      & 22.4 & 1.36 & 16.9 \\
\ourmethod-base    & 10.9 & 1.00 & 13.2 \\
\ourmethod    & \phantom{0}9.8 & 0.87 & 11.6 \\
\bottomrule
\end{tabular}
\ifarxiv\else\vspace{-2mm}\fi
\end{wraptable}
\textbf{Read-time curation produces more compact and effective context than write-time alternatives.}
\Cref{tab:efficiency} reports executor-side token counts and steps. All memory methods inject additional context into the executor prompt, increasing input tokens over the no-memory baseline, but in return the executor solves tasks in fewer steps and generates fewer output tokens. \ourmethod-base achieves this trade-off more favorably: it adds only 1.9K input tokens over no memory, compared to 10.7K for ReasoningBank and 13.4K for \skillos-base, while also reducing executor steps by 18.5\%--21.9\%. RL training further improves all three metrics: \ourmethod reduces input tokens by 10.1\%, output tokens by 13.0\%, and steps by 12.1\% over \ourmethod-base. This is consistent with \citet{shen2026decocted}, who find that higher information density in context correlates with more efficient task completion.\looseness-1

\ifarxiv\else\vspace{-5pt}\fi
\subsection{Ablation Study}
\label{sec:analysis}
\ifarxiv\else\vspace{-3pt}\fi

\begin{figure}[!th]
    \ifarxiv\else\vspace{-3mm}\fi
    \begin{subfigure}{0.49\linewidth}
    \includegraphics[width=\linewidth]{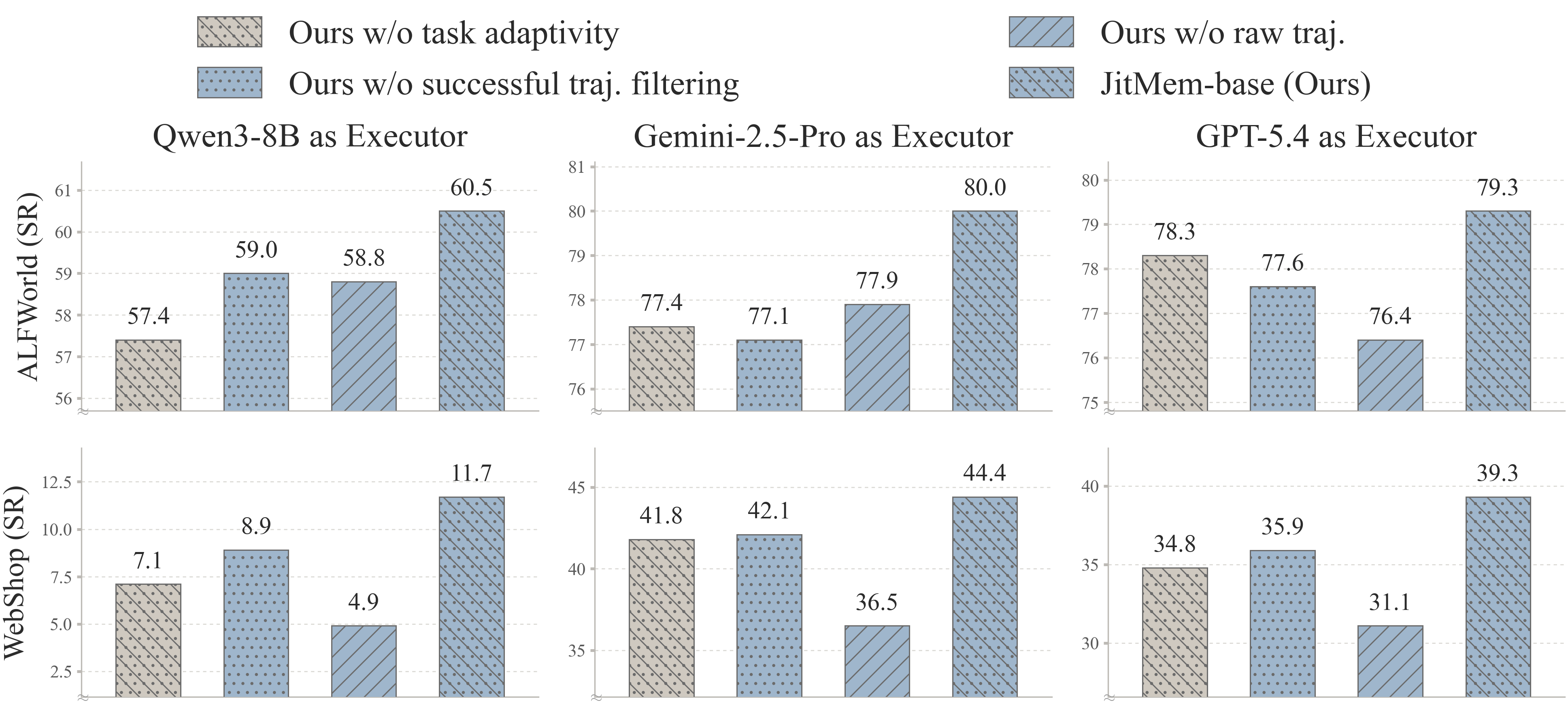} 
    \caption{Ablation of \ourmethod-base}
    \label{fig:baseline-ablation}
    \end{subfigure} 
    \begin{subfigure}{0.49\linewidth}
    \centering
    \includegraphics[width=\linewidth]{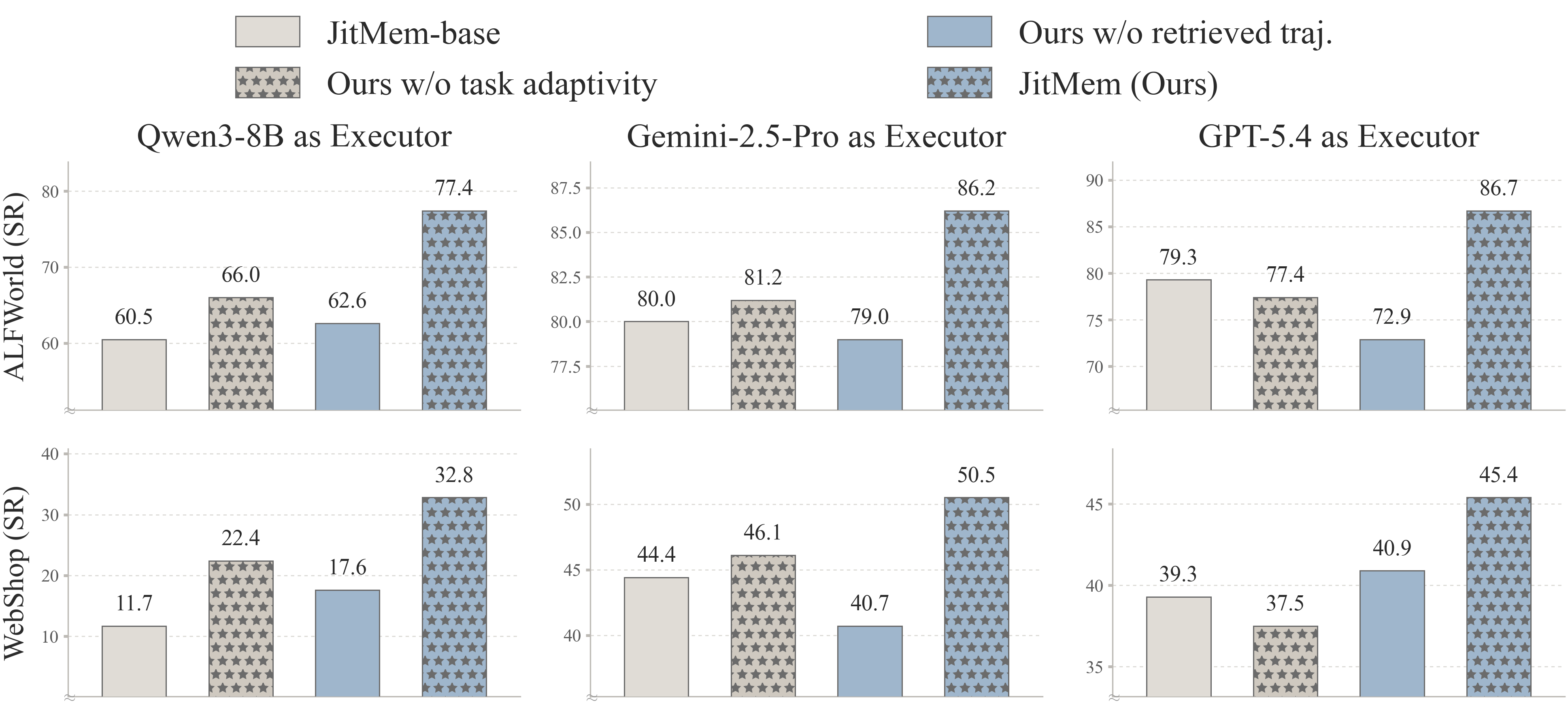} 
    \caption{Ablation of \ourmethod}
    \label{fig:baseline-ablation-w-rl}
    \end{subfigure}
    \ifarxiv\else\vspace{-2mm}\fi
    \caption{Ablation of (a) training-free \ourmethod-base and (b) RL-trained \ourmethod on ALFWorld (top) and WebShop (bottom) across three executors. Each ablation bar removes one design choice, and the rightmost bar is the full method.}
\end{figure}

Within the read-time curation framework, \ourmethod makes several design choices: conditioning on the current task, filtering the memory bank to successful trajectories, and storing raw traces rather than write-time distillations. We ablate each by removing or replacing one at a time, first in the training-free \ourmethod-base (\Cref{fig:baseline-ablation}) and then in the RL-trained \ourmethod (\Cref{fig:baseline-ablation-w-rl}).

\textbf{Task-adaptive conditioning provides gains beyond generic summarization.}
We remove the current task description $x_t$ from the curator's input, reducing it to a query-independent summarizer over retrieved trajectories (``w/o task adaptivity'' in \Cref{fig:baseline-ablation,fig:baseline-ablation-w-rl}). Without RL training, this lowers \ourmethod-base by up to 3.1 on ALFWorld and 4.6 on WebShop across executors. The gap widens after RL training: \ourmethod without task conditioning drops by up to 11.4 on ALFWorld and 10.4 on WebShop, indicating that RL specifically learns to exploit the task signal rather than to compress trajectories more effectively.

\textbf{Quality-filtered storage outperforms label-annotated full storage.}
\ourmethod variants store only trajectories judged as successful. We ablate this by storing all trajectories regardless of outcome and annotating each with its judged correctness label, the practice used in ReasoningBank and \skillos. This drops \ourmethod-base by 1.5--2.9 on ALFWorld and 2.3--3.4 on WebShop across executors. Even with access to correctness annotations, the curator cannot fully suppress the noise introduced by failed trajectories. This result shows that filtering at storage time provides a cleaner retrieval signal.

\textbf{Write-time distillation discards information the curator needs.}
We ablate raw trajectory storage by applying ReasoningBank-style distillation to each trajectory before storage, saving only the distilled items in place of the raw traces. This drops \ourmethod-base by 1.7--2.9 on ALFWorld and 6.8--8.2 on WebShop across executors. Because write-time distillation commits to a query-independent summary, information irreversibly lost at storage time cannot be recovered by the curator at read time (distillation prompt in Appendix~\ref{sec:appendix_hypers}).

\begin{wraptable}{r}{0.42\textwidth}
\centering
\ifarxiv\else\vspace{-4mm}\fi
\setlength{\tabcolsep}{4pt}
\scriptsize
\caption{Staged bank refresh and test bank warm-starting on WebShop.}
\label{tab:iter2}
\vspace{-10pt}
\begin{tabular}{lcc}
    \toprule
    \textbf{Executor}  & \textbf{Score} & \textbf{SR} \\
    \midrule
    \textit{Qwen3-8B} & & \\
    \quad \ourmethod{}            & \valstd{61.1}{0.9} & \valstd{32.8}{1.7} \\
    \quad w/ staged bank refresh  & \valstd{61.7}{0.9} & \valstd{35.6}{0.3} \\
    \quad w/ test bank warm-starting           & \valstd{60.9}{1.0} & \valstd{32.5}{1.5} \\
    \midrule
    \textit{Gemini-2.5-Pro} & & \\
    \quad \ourmethod{}            & \valstd{61.0}{0.8} & \valstd{50.5}{0.8} \\
    \quad w/ staged bank refresh  & \valstd{61.0}{0.3} & \valstd{50.5}{1.1} \\
    \quad w/ test bank warm-starting           & \valstd{61.7}{0.4} & \valstd{50.9}{0.3} \\
    \midrule
    \textit{GPT-5.4} & & \\
    \quad \ourmethod{}            & \valstd{53.8}{0.3} & \valstd{45.4}{0.0} \\
    \quad w/ staged bank refresh  & \valstd{53.6}{0.4} & \valstd{46.3}{0.1} \\
    \quad w/ test bank warm-starting           & \valstd{51.9}{0.3} & \valstd{44.1}{0.4} \\
    \bottomrule
\end{tabular}
\vspace{-10pt}
\end{wraptable}

\textbf{RL learns to distill retrieved experience.}
Does the RL-trained curator genuinely learn to distill retrieved experience, or does it simply learn to generate useful hints from parametric knowledge? We test this by forcing the retriever to return an empty set (``w/o retrieved traj.'' in \Cref{fig:baseline-ablation-w-rl}). Without retrieved context, \ourmethod degrades to or even below the untrained \ourmethod-base across all executors, with SR dropping by up to 14.8 on ALFWorld and 15.2 on WebShop. This confirms that the gains from RL training are grounded in learning how to distill retrieved experience.

\textbf{Staged bank refresh yields modest gains at additional training cost.}
As noted in Section~\ref{sec:method}, the static training bank creates a mild train/test distribution shift. To close this gap, after 100 GRPO steps we discard the original training bank and rebuild it by re-running the executor with the trained curator, then hold the refreshed bank fixed and continue training for 50 additional steps. As shown in \Cref{tab:iter2}, this improves SR by 2.8 for Qwen3-8B and 0.9 for GPT-5.4, while Gemini-2.5-Pro sees no change. The gains are modest relative to the additional training cost, suggesting that the static bank already provides a sufficient training signal.

\textbf{Warm-starting the test bank with training samples provides negligible benefit.}
By default, \ourmethod starts with an empty test bank that grows organically as tasks are solved. We test whether pre-populating the bank with 100 trajectories from the training set improves performance (``w/ test bank warm-starting'' in \Cref{tab:iter2}). The differences are negligible across all three executors, with SR changing by at most 1.3 and remaining within standard deviation. The curator handles sparse retrieval at the beginning of the test sequence gracefully without warm-starting.\looseness-1

\ifarxiv\else\vspace{-5pt}\fi
\subsection{Qualitative Analysis}
\ifarxiv\else\vspace{-3pt}\fi

\begin{figure*}
\centering
    \includegraphics[width=\linewidth]{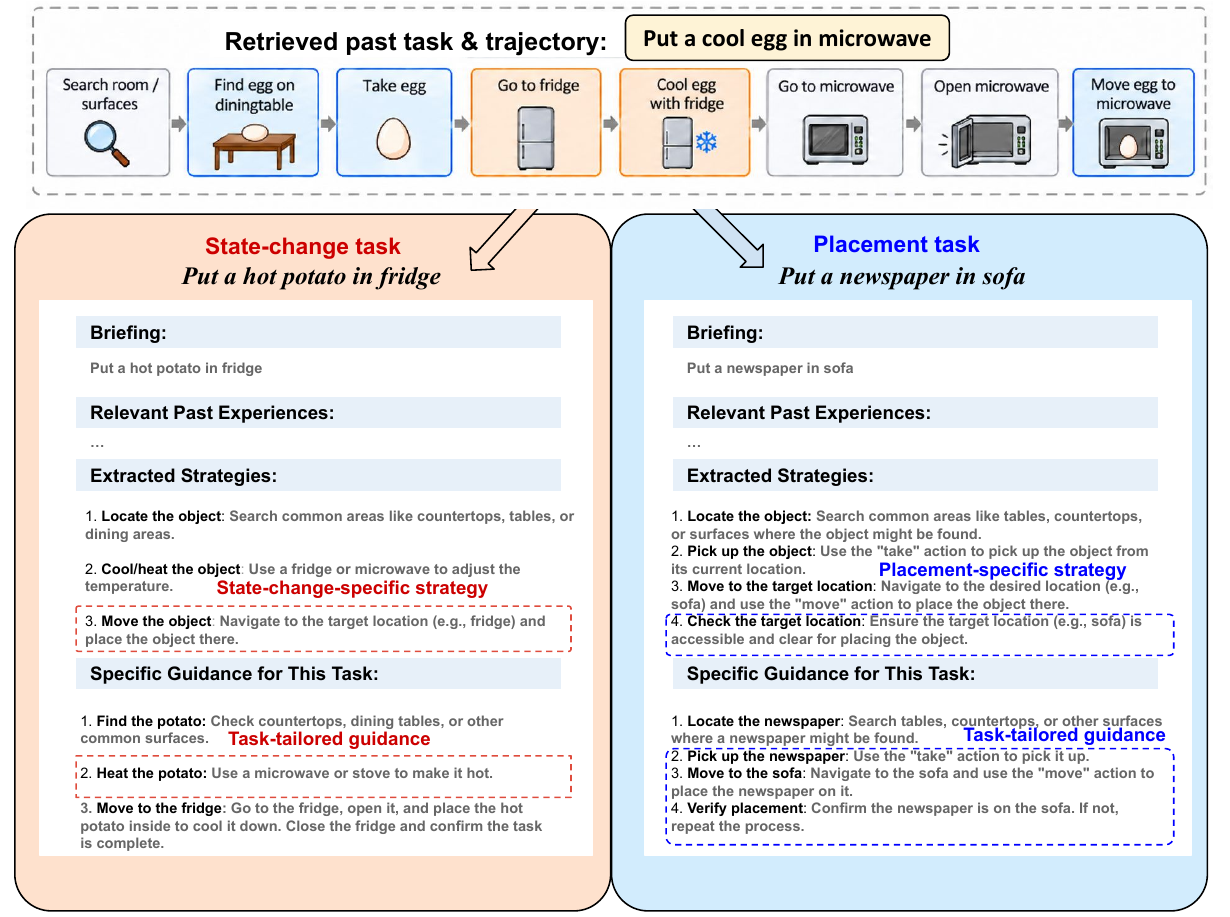}
    \caption{Two tasks retrieve the same past experience but receive different curated payloads. The curator foregrounds state-change guidance for one task and placement-specific guidance for the other.}
    \label{fig:qualitative}
\ifarxiv\else\vspace{-7mm}\fi
\end{figure*}

\textbf{The curator adapts the same experience differently for different tasks.}
Figure~\ref{fig:qualitative} shows two tasks that retrieve the same past experience. For ``put a hot potato in fridge'', the curator foregrounds the state-transition aspect, extracts a heat/cool strategy, and specializes it into guidance for heating the potato before placement. For ``put a newspaper in sofa'', it instead emphasizes placement-specific considerations such as verifying the target location. A write-time artifact would commit to one framing, but the read-time curation produces both from the same stored trace.

\begin{figure*}
\centering
    \includegraphics[width=\linewidth]{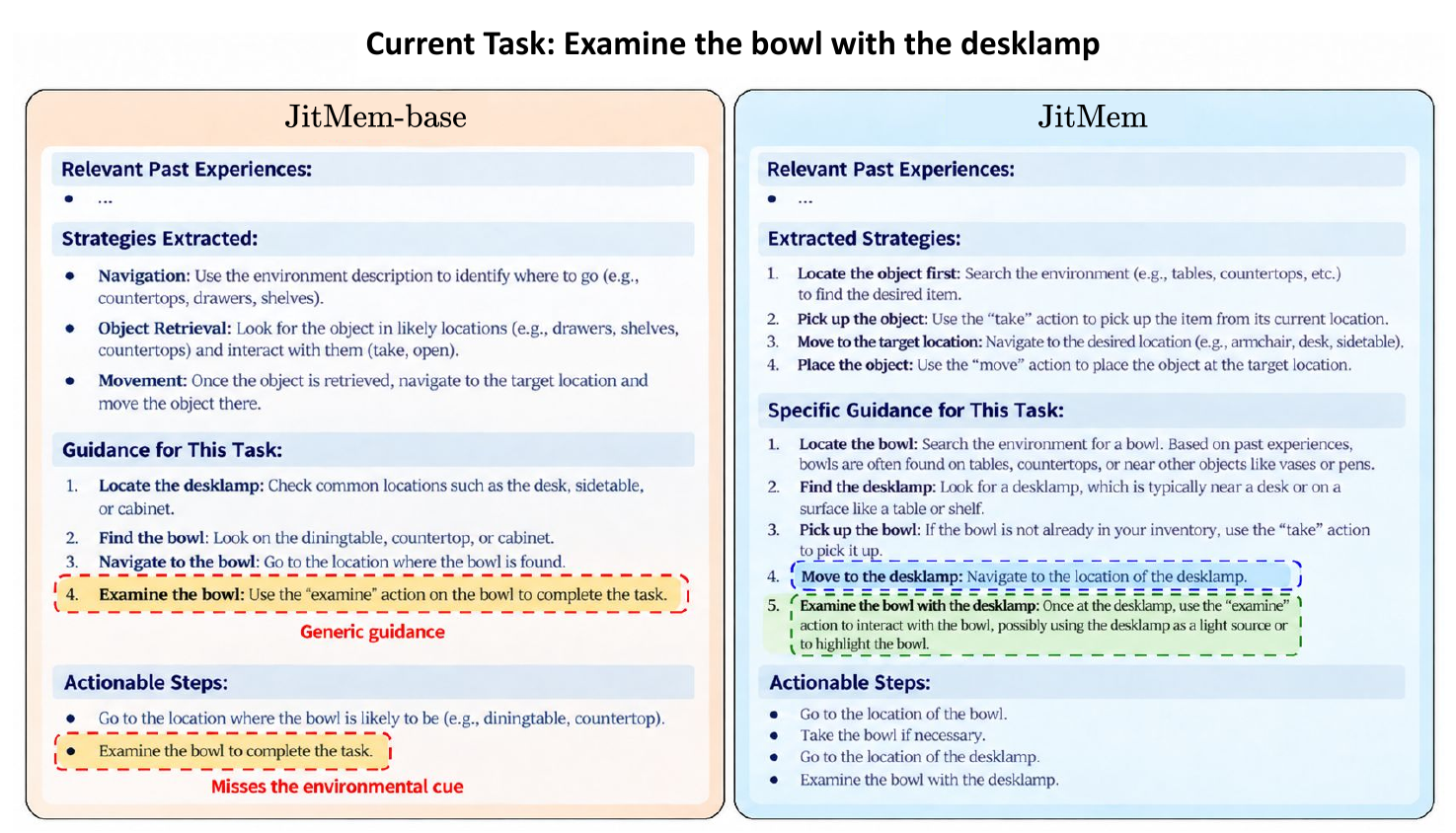}
    \caption{Comparison of untrained vs.\ RL-trained curator payloads on the same input. The untrained curator produces a generic action sequence; the trained curator recovers the environment-specific workflow (move to the desklamp, then examine the bowl with it).}
    \label{fig:qualitative_rl}
\ifarxiv\else\vspace{-6mm}\fi
\end{figure*}

\textbf{RL training induces environment-specific procedural semantics.}
We compare payloads produced by \ourmethod-base and \ourmethod on identical inputs (Figure~\ref{fig:qualitative_rl}). RL adds environment-specific procedural guidance absent from the untrained curator, rather than merely changing the wording. Since such environment-specific workflows are not specified in the curation prompt, their emergence under optimization of $r^{\text{task}}$ suggests that immediate task reward encourages task-relevant curation.

\ifarxiv\else\vspace{-5pt}\fi
\section{Conclusion}
\ifarxiv\else\vspace{-5pt}\fi

We introduced \ourmethod, a memory framework that separates storage from curation by preserving raw trajectories at write time and synthesizing task-adaptive payloads only at read time, once the current task is known. This shift avoids committing prematurely to a single abstraction of past experience and turns curator learning from a delayed future-utility problem into an immediate single-step objective. Across ALFWorld, WebShop, and $\tau^2$-bench, even the untrained read-time curator is competitive with or outperforms strong write-time memory baselines, while RL training further improves effectiveness, efficiency, and transfer across executor models. Ablations further show that RL training learns to distill retrieved experience rather than acquiring standalone task-solving knowledge. Together, these results suggest that effective agent memory depends not only on what experience is stored, but on when and for which task that experience is curated.

\ourmethod has several limitations. The retriever (BM25) is simple and may become a bottleneck as the memory bank grows large and diverse. The curator adds an extra LLM call per task. The payload format is fixed and hand-designed per benchmark. Future work could jointly optimize the payload format, explore stronger retrievers, and extend curation from once per task to turn- or step-level adaptation as new observations arrive.

\newpage
\bibliography{iclr2027_conference}
\bibliographystyle{salesforce}

\appendix
\newpage

\section{Experimental Details and Hyperparameters}
\label{sec:appendix_hypers}

\begin{tcolorbox}[breakable,enhanced, left=\promptleft, right=\promptright, top=2pt, bottom=2pt, enlarge top by=0.1cm, enlarge bottom by=0.1cm, title={\hspace{\ifarxiv 2mm\else 1cm\fi} \ourmethod Prompt (ALFWorld)}, fonttitle=\bfseries\small]
\begin{quote}
\begin{lstlisting}[basicstyle=\ttfamily\scriptsize]
<System Prompt>
You are a Memory Curator. You will be given a task that an AI agent needs to solve in a household (ALFWorld) environment, along with retrieved past experiences from similar successful tasks.

Your job: synthesize these raw memories into a concise, actionable briefing that will help the agent solve the current task.

Each memory contains:
- A past task and the trajectory that solved it

Your output should:
1. Identify which past experiences are most relevant
2. Extract strategies that worked on similar tasks (e.g., where to find objects, useful action orders)
3. Give specific guidance for THIS task

Be concise - the agent has limited context.

<User Prompt>
Question: {query}

### Retrieved Memories:
Memory 1:
Question: {memory1_query}
Trajectory:
{memory1_trajectory}
...
\end{lstlisting}
\end{quote}
\end{tcolorbox}

\begin{tcolorbox}[breakable,enhanced, left=\promptleft, right=\promptright, top=2pt, bottom=2pt, enlarge top by=0.1cm, enlarge bottom by=0.1cm, title={\hspace{\ifarxiv 2mm\else 1cm\fi} \ourmethod Prompt (WebShop)}, fonttitle=\bfseries\small]
\begin{quote}
\begin{lstlisting}[basicstyle=\ttfamily\scriptsize]
<System Prompt>
You are a Memory Curator. You will be given a shopping task that an AI agent needs to solve in the WebShop e-commerce environment, along with retrieved past experiences from similar shopping tasks the agent completed with partial or full success.

Your job: synthesize these raw memories into a concise, actionable briefing that will help the agent buy the correct item for the current task.

Each memory contains:
- A past shopping instruction and the trajectory (search queries, product clicks, option selections, and the final purchase) that addressed it

Your output should:
1. Identify which past experiences target similar products and attributes
2. Extract strategies that worked - how the search was phrased, how the right product was chosen, how the requested options (e.g. color, size) were set, and how the price constraint was met before purchasing. A past trajectory may have only partially satisfied its instruction, so keep what generalizes.
3. Give specific, actionable guidance for THIS task

Do not assume a past trajectory's exact product is still available, and do not rely on specific product IDs.
Be concise - the agent has limited context.

<User Prompt>
Question: {query}

### Retrieved Memories:
Memory 1:
Question: {memory1_query}
Trajectory:
{memory1_trajectory}
...
\end{lstlisting}
\end{quote}
\end{tcolorbox}

\begin{tcolorbox}[breakable,enhanced, left=\promptleft, right=\promptright, top=2pt, bottom=2pt, enlarge top by=0.1cm, enlarge bottom by=0.1cm, title={\hspace{\ifarxiv 2mm\else 1cm\fi} \ourmethod Prompt ($\tau^2\text{-bench}$)}, fonttitle=\bfseries\small]
\begin{quote}
\begin{lstlisting}[basicstyle=\ttfamily\scriptsize]
<System Prompt>
You are a Memory Curator. You will be given a customer-service request that an AI agent must handle by talking to the user and calling tools, along with retrieved past experiences from similar requests the agent resolved successfully.

Your job: synthesize these raw memories into a concise, actionable briefing that helps the agent resolve the current request while following the domain policy.

Each memory contains:
- A past customer request and the transcript that resolved it. In the transcript, [USER] lines are the customer, [AGENT] lines are the agent's messages or its tool calls written as fn(arg=value), and [TOOL] lines are the tool results.

Your output should:
1. Identify which past experiences target the most similar request type
2. Extract the resolution strategy that worked, described as a short ordered plan: the read/lookup tools used to gather state, the confirmation the agent obtained from the user, the policy conditions verified, and the write/mutating tool call(s) that completed the task
3. Give specific, actionable guidance for THIS request

Do not copy concrete identifiers or values from past memories (reservation IDs, confirmation numbers, user IDs, flight numbers, prices, dates) - always look up the current case with the tools. Never suggest an action that conflicts with the domain policy.
Be concise - the agent has limited context.

<User Prompt>
Question: {query}

### Retrieved Memories:
Memory 1:
Question: {memory1_query}
Trajectory:
{memory1_trajectory}
...
\end{lstlisting}
\end{quote}
\end{tcolorbox}

\begin{tcolorbox}[breakable,enhanced, left=\promptleft, right=\promptright, top=2pt, bottom=2pt, enlarge top by=0.1cm, enlarge bottom by=0.1cm, title={\hspace{\ifarxiv 2mm\else 1cm\fi} Executor Prompt (ALFWorld)}, fonttitle=\bfseries\small]
\label{pt:alfworld_exec}
\begin{quote}
\begin{lstlisting}[basicstyle=\ttfamily\scriptsize]
You are an expert agent operating in the ALFRED Embodied Environment. Your task is to: {task_description}

Here are past experiences and trajectories that might be helpful for your decision:

{retrieved_context}

## Current Progress

Prior to this step, you have already taken {step_count} step(s). Below are the most recent {history_length} observations and the corresponding actions you took: {action_history}
You are now at step {current_step} and your current observation is: {current_observation}
Your admissible actions of the current situation are: [{admissible_actions}].

Now it's your turn to take an action.
You should first reason step-by-step about the current situation with the help of past relevant experiences.
Once you've finished your reasoning, you should choose an admissible action for current step and MUST present it within <action> </action> tags.
\end{lstlisting}
\end{quote}
\end{tcolorbox}

\begin{tcolorbox}[breakable,enhanced, left=\promptleft, right=\promptright, top=2pt, bottom=2pt, enlarge top by=0.1cm, enlarge bottom by=0.1cm, title={\hspace{\ifarxiv 2mm\else 1cm\fi} Executor Prompt (WebShop)}, fonttitle=\bfseries\small]
\begin{quote}
\begin{lstlisting}[basicstyle=\ttfamily\scriptsize]
You are an expert agent operating in the WebShop e-commerce environment. 
Your task is to: {task_description}.

Here are past experiences and trajectories that might be helpful for your decision:

{retrieved_context}

## Current Progress

Prior to this step, you have already taken {step_count} step(s). Below are the most recent {history_length} observations and the corresponding actions you took: {action_history}
You are now at step {current_step} and your current observation is: {current_observation}.
Your admissible actions of the current situation are:
[
{available_actions}
].

Now it's your turn to take one action for the current step.
You should first reason step-by-step about the current situation with the help of past relevant experiences, then think carefully which admissible action best advances the shopping goal.
Once you've finished your reasoning, you should choose an admissible action for current step and present it within <action> </action> tags.

WebShop search guidance:
- Use search[<your query>] with a short core product query, such as the product type or category.
- Do not put color, size, price, or every requested attribute into search[<your query>]. Handle those by opening a product page and selecting/clicking options when available.
- If a search returns zero results, retry with a shorter broader product query, not a longer query.
- The goal is to inspect/select a matching product and eventually click[buy now].
\end{lstlisting}
\end{quote}
\end{tcolorbox}

\begin{tcolorbox}[breakable,enhanced, left=\promptleft, right=\promptright, top=2pt, bottom=2pt, enlarge top by=0.1cm, enlarge bottom by=0.1cm, title={\hspace{\ifarxiv 2mm\else 1cm\fi} Executor Prompt ($\tau^2$\text{-bench})}, fonttitle=\bfseries\small]
\begin{quote}
\begin{lstlisting}[basicstyle=\ttfamily\scriptsize]
<instructions>
You are a customer service agent that helps the user according to the <policy> provided below.
In each turn you can either:
- Send a message to the user.
- Make a tool call.
You cannot do both at the same time.
Try to be helpful and always follow the policy. Always make sure you generate valid JSON only.
</instructions>

<policy>
[Content Omitted]
</policy>

Here are past experiences and trajectories that might be helpful for your decision. You can use it when you feel it's relevant. At each step, first reason about the current situation with the help of past relevant experiences, including explicitly discuss if you want to use past experiences or not, and then take action.

{retrieved_context}

\end{lstlisting}
\end{quote}
\end{tcolorbox}

\begin{tcolorbox}[breakable,enhanced, left=\promptleft, right=\promptright, top=2pt, bottom=2pt, enlarge top by=0.1cm, enlarge bottom by=0.1cm, title={\hspace{\ifarxiv 2mm\else 1cm\fi} LLM-as-Judge Prompt ($\tau^2$-bench)}, fonttitle=\bfseries\small]
\begin{quote}
\begin{lstlisting}[basicstyle=\ttfamily\scriptsize]
You are an expert judge evaluating whether a customer-service agent successfully satisfied a customer's request. Output a single JSON object and nothing else.
# Task
You will be given (1) the customer's request and (2) the full conversation between the agent and the customer. The conversation contains [USER] lines (the customer), [AGENT] lines (the agent's messages, or its tool calls written as fn(arg=value)), and [TOOL] lines (the tool results). Determine whether the agent fully satisfied the customer's request.
## What "success" means
- The agent must have actually carried out what the customer asked - the correct action (e.g. booking, cancellation, update, refund, troubleshooting fix) must be completed via the appropriate tool call, and the tool result must confirm it succeeded.
- Credit only outcomes that the [TOOL] results confirm. Do not credit effects the agent merely stated, promised, or planned. Ignore the agent's own claims of completion; rely on the tool results.
- If the request required following a process (verifying identity/eligibility, confirming before an irreversible action), that process must be evidenced in the transcript.
## Strictness
- If the transcript is ambiguous about whether the request was fully satisfied, output success=false.
- Partial completion is failure: either the customer's request is fully satisfied or the conversation is a failure.
- A conversation that ends by giving up, escalating to a human, or hitting the step limit without completing the request is a failure.
# Output
Output exactly one JSON object with these fields and nothing else:
{ "success": <true|false>, "reasoning": "<one or two sentences citing the specific tool results that prove success or failure>" }
\end{lstlisting}
\end{quote}
\end{tcolorbox}

\begin{tcolorbox}[breakable,enhanced, left=\promptleft, right=\promptright, top=2pt, bottom=2pt, enlarge top by=0.1cm, enlarge bottom by=0.1cm, title={\hspace{\ifarxiv 2mm\else 1cm\fi} ReasoningBank-style Distillation Prompt for Ablation Study (ALFWorld)}, fonttitle=\bfseries\small]
\begin{quote}
\begin{lstlisting}[basicstyle=\ttfamily\scriptsize]
You are an expert in household task planning. You will be given a task and a trajectory representing how an agent successfully completed the task in a household environment.

## Guidelines
Extract and summarize useful insights as memory items that would help an agent solve similar household tasks in the future.

## Important notes
  - Think about why the trajectory succeeded, then summarize the insights.
  - Extract *at most 3* memory items.
  - Do not repeat similar or overlapping items.
  - Focus on generalizable strategies (e.g., where to find objects, what order to do actions), not specific object names or locations.

## Output Format
Your output must strictly follow this Markdown format:

```
# Memory Item i
## Title <short title>
## Description <one sentence summary>
## Content <1-3 sentences of actionable insight>
```
\end{lstlisting}
\end{quote}
\end{tcolorbox}

\begin{tcolorbox}[breakable,enhanced, left=\promptleft, right=\promptright, top=2pt, bottom=2pt, enlarge top by=0.1cm, enlarge bottom by=0.1cm, title={\hspace{\ifarxiv 2mm\else 1cm\fi} ReasoningBank-style Distillation Prompt for Ablation Study (WebShop)}, fonttitle=\bfseries\small]
\begin{quote}
\begin{lstlisting}[basicstyle=\ttfamily\scriptsize]
You are an expert in online-shopping task planning. You will be given a task and a trajectory representing how an agent successfully completed a shopping task on a web store.

## Guidelines
Extract and summarize useful insights as memory items that would help an agent solve similar shopping tasks in the future.

## Important notes
  - Think about why the trajectory succeeded, then summarize the insights.
  - Extract *at most 3* memory items.
  - Do not repeat similar or overlapping items.
  - Focus on generalizable strategies, not specific product IDs, prices, or properties.

## Output Format
Your output must strictly follow this Markdown format:

```
# Memory Item i
## Title <short title>
## Description <one sentence summary>
## Content <1-3 sentences of actionable insight>
```
\end{lstlisting}
\end{quote}
\end{tcolorbox}

\FloatBarrier
\noindent \textbf{Prompts.} All curator, executor, LLM-as-judge, and distillation prompts are provided in Appendix~\ref{sec:appendix_hypers}. Our executor and judge prompts for ALFWorld and WebShop follow \skillos~\citep{ouyang2026skillos}. For $\tau^2$-bench~\citep{barres2025tau}, we use the executor prompt from the official benchmark and design a separate judge prompt. The ReasoningBank-style distillation prompt used in the ablation study (\Cref{sec:analysis}) is also included.

\begin{table}[h]
\centering
\caption{No-memory baseline reproduction. ``Reported'' from \citet{ouyang2026skillos}; ``Reproduced'' from our runs. In all cases, our reproduced performance is at or below the reported results, ensuring gains are measured conservatively.}
\label{tab:reproduction}
\begin{tabular}{lccc}
    \toprule
    \multirow{2}{*}{\textbf{Source}}
    & \textbf{ALFWorld}
    & \multicolumn{2}{c}{\textbf{WebShop}} \\
    \cmidrule(lr){2-2} \cmidrule(lr){3-4}
    & \textbf{SR} & \textbf{Score} & \textbf{SR} \\
    \midrule
    \multicolumn{4}{c}{\cellcolor{gray!15}\textit{Executor: Qwen3-8B}} \\
    \skillos Reported   & \valstd{47.9}{1.2} & \valstd{33.3}{0.7} & \valstd{9.8}{0.5} \\
    Reproduced (non-thinking) & \valstd{34.5}{0.3} & \valstd{36.4}{0.2} & \valstd{8.6}{1.1} \\
    Reproduced (thinking) & \valstd{42.1}{1.0} & \valstd{29.0}{0.3} & \valstd{4.4}{0.7} \\
    \midrule 
    \multicolumn{4}{c}{\cellcolor{gray!15}\textit{Executor: Gemini-2.5-Pro}} \\
    \skillos Reported   & \valstd{66.4}{2.0} & \valstd{48.6}{0.3} & \valstd{38.4}{0.5} \\
    Reproduced & \valstd{63.6}{1.3} & \valstd{47.3}{0.8} & \valstd{37.5}{0.5} \\
    \bottomrule
\end{tabular}
\end{table}

\noindent \textbf{Reproducing baselines.}
Since \skillos~\citep{ouyang2026skillos} does not fully specify its inference settings (e.g., thinking vs.\ non-thinking mode for Qwen3-8B), we ran preliminary experiments to identify configurations that reproduce their reported no-memory baselines (\Cref{tab:reproduction}). The closest match is thinking mode on ALFWorld and non-thinking mode on WebShop. We remove the default thinking-format instruction from Qwen3-8B executor prompts, as it causes degenerate outputs in non-thinking mode (see the executor prompt in Appendix~\ref{pt:alfworld_exec}). On WebShop, we could not fully reproduce the reported no-memory result with this change alone, so we add a short search-guidance paragraph to the Qwen3-8B executor prompt, inspired by the WebShop evaluation script.\footnote{\url{https://huggingface.co/datasets/zhangdw/webshop/blob/main/evaluate.py}} As \Cref{tab:reproduction} shows, we calibrated this guidance so that our no-memory baseline remains at or below the reported result (8.6 vs.\ 9.8 SR), ensuring our gains are measured conservatively. For Gemini-2.5-Pro and GPT-5.4, we reproduce the reported baselines without modification.

\noindent \textbf{Hyperparameters.}
Based on the reproduction study above, we adopt the following settings.

\emph{Executor.} All three executors (Qwen3-8B, Gemini-2.5-Pro, GPT-5.4) use temperature 1.0 and maximum output length 4096. For Qwen3-8B, we use thinking mode on ALFWorld and non-thinking mode on WebShop. Search guidance is added to the Qwen3-8B WebShop executor prompt only. We set a history window of 3 and a maximum of 30 interaction turns for ALFWorld and WebShop, and 200 turns for $\tau^2$-bench.

\emph{Curator.} For GPT-5.4 and Gemini-2.5-Pro curators, we use temperature 1.0. For both untrained and trained Qwen3-8B curators (non-thinking), we use temperature 0.6 with top-$p$ 0.95 and top-$k$ 20.

\emph{Retrieval and batching.} We use $k{=}3$ retrieved trajectories for all main results; an ablation over $k \in \{3,5\}$ is provided in Appendix~\ref{sec:appendix_results}. The evaluation batch size is 10 for ALFWorld and WebShop and 5 for $\tau^2$-bench.

\emph{Training.} Full GRPO training hyperparameters are reported in \Cref{tab:rl-hparams}. During curator training, the Qwen3-8B executor runs in non-thinking mode for efficiency. A single training run takes approximately 21 hours on ALFWorld and 27 hours on WebShop.

\emph{Infrastructure.} All experiments run on a single server with 8$\times$NVIDIA H200 GPUs, 2$\times$Intel Xeon Platinum 8488C (96 logical cores), and 2\,TB system RAM. We serve Qwen3-8B with \textbf{vLLM}~\citep{kwon2023efficient} using tensor parallelism $\text{TP}{=}1$, data parallelism $\text{DP}{=}4$, and a maximum model length of 40960.\looseness-1

\begin{table}[h]
  \centering
  \caption{RL (GRPO) training hyperparameters.}
  \label{tab:rl-hparams}
  \begin{tabular}{@{}ll@{}}
    \toprule
    \textbf{Hyperparameter} & \textbf{Value} \\
    \midrule
    \multicolumn{2}{@{}l}{\emph{Optimization}}\\
    Base policy (curator)             & Qwen3-8B (non-thinking) \\
    Advantage estimator               & GRPO (no std.\ normalization) \\
    Learning rate                     & $1\times10^{-6}$ \\
    LR schedule                       & constant with warmup \\
    Warmup steps                      & $5$ (ratio $0.05$ of $100$ steps) \\
    Train batch size (prompts)        & $32$ \\
    Policy mini-batch size            & $32$ \\
    Max RL steps                      & $100$ \\
    \midrule
    \multicolumn{2}{@{}l}{\emph{Objective}}\\
    KL in reward                      & disabled \\
    KL loss                           & enabled, low-variance KL, coef.\ $1\times10^{-3}$ \\
    Clip range (low, high)            & $(0.2,\ 0.2)$ \\
    Loss aggregation                  & token-mean \\
    \midrule
    \multicolumn{2}{@{}l}{\emph{Rollout / generation (curator)}}\\
    Samples per prompt (group size)   & $8$ \\
    Max prompt length                 & $32768$ \\
    Max response length               & $8192$ (ALFWorld) / $4096$ (WebShop) \\
    Sampling temperature (train)      & $1.0$ \\
    \midrule
    \multicolumn{2}{@{}l}{\emph{Executor / environment}}\\
    Executor model                    & Qwen3-8B (non-thinking), served via vLLM \\
    Executor temperature / top-$p$ / top-$k$ & $1.0$ / $0.95$ / $20$ \\
    Executor max new tokens           & $4096$ \\
    Max env.\ steps per game          & $30$ \\
    Retrieved memories per task       & $3$ \\
    \bottomrule
  \end{tabular}
\end{table}

\FloatBarrier
\section{Additional Results}
\label{sec:appendix_results}

\textbf{Sensitivity to retrieval count $k$.} \Cref{tab:ablation-hyper} compares $k{=}3$ and $k{=}5$ retrieved trajectories on WebShop across all three executors. Performance is stable across both settings: SR varies by less than 2 points in all cases, and the differences remain within standard deviation for Qwen3-8B and GPT-5.4. This indicates that \ourmethod is not sensitive to the retrieval count, and we use $k{=}3$ for all main results for efficiency.

\begin{table}[!h]
\centering
\caption{Ablation of the number of retrieved trajectories $k$ on WebShop.}
\label{tab:ablation-hyper}
\begin{tabular}{lcc}
    \toprule
    \textbf{Executor}  & \textbf{Score} & \textbf{SR} \\
    \midrule
    \cellcolor{gray!15}\textit{Qwen3-8B} & \cellcolor{gray!15} & \cellcolor{gray!15} \\
    \quad $k = 3$             & \valstd{61.1}{0.9} & \valstd{32.8}{1.7} \\
    \quad $k = 5$  &    \valstd{60.5}{0.4} & \valstd{32.8}{0.3} \\
    \midrule
    \cellcolor{gray!15} \textit{Gemini-2.5-Pro} & \cellcolor{gray!15}&\cellcolor{gray!15} \\
    \quad $k = 3$            & \valstd{61.0}{0.8} & \valstd{50.5}{0.8} \\
    \quad $k = 5$           & \valstd{59.2}{0.4} & \valstd{48.6}{0.7} \\
    \midrule
    \cellcolor{gray!15} \textit{GPT-5.4} & \cellcolor{gray!15}&\cellcolor{gray!15} \\
    \quad $k = 3$            & \valstd{53.8}{0.3} & \valstd{45.4}{0.0} \\
    \quad $k = 5$           & \valstd{53.4}{0.7} & \valstd{45.3}{0.8} \\
    \bottomrule
\end{tabular}
\end{table}

\textbf{Consolidated ablation results.} \Cref{tab:ablation_study} collects all ablation and baseline results from the main text into a single table for easy comparison. For each executor, the table first lists baselines, then \ourmethod-base ablations, and finally \ourmethod ablations. 

The \ourmethod-base ablations isolate individual design choices:
\begin{itemize}[leftmargin=*,noitemsep,topsep=2pt]
\item \emph{w/o task adaptivity}: removes current task description $x_t$ from the curator's input, reducing it to a query-independent summarizer.
\item \emph{w/o successful traj.\ filtering}: stores all trajectories with correctness labels instead of filtering to successful ones.
\item \emph{w/o raw traj.}: applies ReasoningBank-style distillation at write time and stores only the distilled items.
\end{itemize}

The \ourmethod ablations probe what RL training learns:
\begin{itemize}[leftmargin=*,noitemsep,topsep=2pt]
\item \emph{w/o retrieved traj.}: forces the retriever to return an empty set, isolating parametric knowledge from episodic retrieval.
\item \emph{w/o task adaptivity}: removes current task description $x_t$ from the RL-trained curator.
\item \emph{w/ staged bank refresh}: rebuilds the training bank with curator-augmented trajectories after 100 GRPO steps and continues training for 50 more.
\item \emph{w/ test bank warm-starting}: pre-populates the test bank with 100 training-set trajectories to mitigate the cold-start effect.
\end{itemize}

Key findings from \Cref{tab:ablation_study}:
\begin{enumerate}[leftmargin=*,noitemsep,topsep=2pt]
\item \textbf{Every design choice contributes.} Every \ourmethod-base ablation degrades performance, confirming that task-adaptive conditioning, quality-filtered storage, and raw trajectory retention each contribute independently. The largest drop comes from removing raw trajectories on WebShop (up to $-8.2$ SR), underscoring that write-time distillation discards information the curator needs.
\item \textbf{RL learns to distill retrieved experience.} Removing retrieved trajectories from the RL-trained \ourmethod causes the largest degradation (up to $-14.8$ SR on ALFWorld and $-15.2$ SR on WebShop). Removing task adaptivity also degrades substantially, with the gap widening compared to the untrained variant.
\item \textbf{Default training and evaluation settings are sufficient.} Staged bank refresh provides modest gains ($+2.8$ SR at best), and test bank warm-starting has negligible effect, suggesting the static training bank and empty-start evaluation already work well.
\end{enumerate}

\begin{table*}[t]
\centering
\setlength{\tabcolsep}{4pt}
\small
\setlength{\belowcaptionskip}{6.0pt}
\caption{Consolidated ablation and baseline results on ALFWorld and WebShop across three executors. Each block reports baselines, \ourmethod-base ablations (\textcolor{blue}{blue}), and \ourmethod ablations (\textcolor{blue}{blue}). \frozenmark~denotes prompted curator; \firemark~denotes RL-trained. Mean $\pm$ std over 3 runs.}
\label{tab:ablation_study}
\resizebox{\ifarxiv 0.7\textwidth \else 0.88\textwidth \fi}{!}{
\begin{tabular}{llccc}
    \toprule
    \multirow{2}{*}{\textbf{Methods}}
    & \textbf{Curator}
    & \textbf{ALFWorld}
    & \multicolumn{2}{c}{\textbf{WebShop}} \\
    \cmidrule(lr){3-3} \cmidrule(lr){4-5}
    & 
    & \textbf{SR} & \textbf{Score} & \textbf{SR} \\
    \midrule
    \multicolumn{5}{c}{\cellcolor{gray!15}\textit{Executor: Qwen3-8B}} \\
    No Memory & ---
    & \valstd{47.9}{1.2} & \valstd{33.3}{0.7} & \valstd{\phantom{0}9.8}{0.5} \\
    ReasoningBank & \adjustbox{valign=c}{\frozenmark} Qwen3-8B
    & \valstd{55.7}{3.1} & \valstd{35.4}{1.1} & \valstd{11.4}{0.9} \\
    MemP & \adjustbox{valign=c}{\frozenmark} Qwen3-8B
    & \valstd{49.7}{0.7} & \valstd{35.7}{0.9} & \valstd{12.0}{0.5} \\
    \skillos{}-base & \adjustbox{valign=c}{\frozenmark} Qwen3-8B
    & \valstd{53.1}{2.5} & \valstd{38.6}{0.9} & \valstd{13.6}{0.8} \\
    \skillos{}-gemini & \adjustbox{valign=c}{\frozenmark} Gemini-2.5-Pro
    & \valstd{50.7}{3.6} & \valstd{38.1}{1.0} & \valstd{13.2}{0.9} \\
    \skillos{} & \adjustbox{valign=c}{\firemark} Qwen3-8B
    & \valstd{61.2}{4.6} & \valstd{40.6}{0.7} & \valstd{16.5}{0.7} \\
    \hdashline
    \rowcolor{blockrow} \ourmethod-base & \adjustbox{valign=c}{\frozenmark} Qwen3-8B
    & \valstd{60.5}{2.6} & \valstd{32.5}{3.2} & \valstd{11.7}{0.5} \\
    \rowcolor{blockrow} \quad \textcolor{blue}{w/o task adaptivity} & \adjustbox{valign=c}{\frozenmark} Qwen3-8B
    & \textcolor{blue}{\valstd{57.4}{2.0}} &  \textcolor{blue}{\valstd{25.3}{0.9}} & \textcolor{blue}{\valstd{\phantom{0}7.1}{0.1}} \\
    \rowcolor{blockrow} \quad \textcolor{blue}{w/o successful traj.\ filtering} & \adjustbox{valign=c}{\frozenmark} Qwen3-8B
    & \textcolor{blue}{\valstd{59.0}{2.2}} &  \textcolor{blue}{\valstd{23.5}{2.9}} &  \textcolor{blue}{\valstd{\phantom{0}8.9}{1.7}} \\ 
    \rowcolor{blockrow} \quad \textcolor{blue}{w/o raw traj.} & \adjustbox{valign=c}{\frozenmark} Qwen3-8B
    & \textcolor{blue}{\valstd{58.8}{1.5}} & \textcolor{blue}{\valstd{17.8}{1.7}} & \textcolor{blue}{\valstd{\phantom{0}4.9}{0.7}}  \\ 
    \rowcolor{blockrow} \ourmethod & \adjustbox{valign=c}{\firemark} Qwen3-8B
    & \valstd{77.4}{2.9} & \valstd{61.1}{0.9} & \valstd{32.8}{1.7} \\
    \rowcolor{blockrow} \quad \textcolor{blue}{w/o retrieved traj.} & \adjustbox{valign=c}{\firemark} Qwen3-8B
    & \textcolor{blue}{\valstd{62.6}{1.2}} &  \textcolor{blue}{\valstd{47.9}{1.6}} & \textcolor{blue}{\valstd{17.6}{1.1}} \\ 
    \rowcolor{blockrow} \quad \textcolor{blue}{w/o task adaptivity} & \adjustbox{valign=c}{\firemark} Qwen3-8B
    & \textcolor{blue}{\valstd{66.0}{4.7}} & \textcolor{blue}{\valstd{49.6}{1.4}} & \textcolor{blue}{\valstd{22.4}{0.4}}  \\ 
    \rowcolor{blockrow} \quad \textcolor{blue}{w/ staged bank refresh} & \adjustbox{valign=c}{\firemark} Qwen3-8B + stage 2
    & -- & \textcolor{blue}{\valstd{61.7}{0.9}} & \textcolor{blue}{\valstd{35.6}{0.3}} \\
    \rowcolor{blockrow} \quad \textcolor{blue}{w/ test bank warm-starting} & \adjustbox{valign=c}{\firemark} Qwen3-8B
    & -- & \textcolor{blue}{\valstd{60.9}{1.0}} & \textcolor{blue}{\valstd{32.5}{1.5}} \\
    \midrule
    \multicolumn{5}{c}{\cellcolor{gray!15}\textit{Executor: Gemini-2.5-Pro}} \\
    No Memory & ---
    & \valstd{66.4}{2.0} & \valstd{48.6}{0.3} & \valstd{38.4}{0.5} \\
    ReasoningBank & \adjustbox{valign=c}{\frozenmark} Qwen3-8B
    & \valstd{71.4}{2.9} & \valstd{47.1}{1.0} & \valstd{38.0}{0.6} \\
    ReasoningBank & \adjustbox{valign=c}{\frozenmark} Gemini-2.5-Pro
    & \valstd{78.6}{2.9} & \valstd{50.8}{1.5} & \valstd{40.2}{1.3} \\
    MemP & \adjustbox{valign=c}{\frozenmark} Qwen3-8B
    & \valstd{74.3}{3.4} & \valstd{51.9}{1.9} & \valstd{40.3}{1.3} \\
    MemP & \adjustbox{valign=c}{\frozenmark} Gemini-2.5-Pro
    & \valstd{77.1}{2.1} & \valstd{51.3}{1.2} & \valstd{39.8}{1.0} \\
    \skillos{}-base & \adjustbox{valign=c}{\frozenmark} Qwen3-8B
    & \valstd{70.7}{3.0} & \valstd{52.8}{1.0} & \valstd{39.6}{0.8} \\
    \skillos{}-gemini & \adjustbox{valign=c}{\frozenmark} Gemini-2.5-Pro
    & \valstd{79.3}{2.6} & \valstd{54.7}{1.0} & \valstd{41.0}{1.2} \\
    \skillos{} & \adjustbox{valign=c}{\firemark} Qwen3-8B
    & \valstd{80.2}{3.1} & \valstd{56.0}{0.7} & \valstd{41.3}{0.8} \\
    \hdashline
    \rowcolor{blockrow} \ourmethod-base  & \adjustbox{valign=c}{\frozenmark} Qwen3-8B
    & \valstd{80.0}{1.5} & \valstd{54.7}{1.4} & \valstd{44.4}{0.6} \\
    \rowcolor{blockrow} \quad \textcolor{blue}{w/o task adaptivity} & \adjustbox{valign=c}{\frozenmark} Qwen3-8B
    & \textcolor{blue}{\valstd{77.4}{2.4}} & \textcolor{blue}{\valstd{51.2}{1.3}} & \textcolor{blue}{\valstd{41.8}{1.4}} \\ 
    \rowcolor{blockrow} \quad \textcolor{blue}{w/o successful traj.\ filtering} & \adjustbox{valign=c}{\frozenmark} Qwen3-8B
    & \textcolor{blue}{\valstd{77.1}{3.1}} & \textcolor{blue}{\valstd{52.0}{1.0}} & \textcolor{blue}{\valstd{42.1}{0.7}}\\
    \rowcolor{blockrow} \quad \textcolor{blue}{w/o raw traj.} & \adjustbox{valign=c}{\frozenmark} Qwen3-8B
    & \textcolor{blue}{\valstd{77.9}{2.7}} & \textcolor{blue}{\valstd{45.5}{1.0}} & \textcolor{blue}{\valstd{36.5}{0.8}} \\
    \rowcolor{blockrow} \ourmethod  & \adjustbox{valign=c}{\firemark} Qwen3-8B
    & \valstd{86.2}{1.9}  & \valstd{61.0}{0.8} & \valstd{50.5}{0.8} \\
    \rowcolor{blockrow} \quad \textcolor{blue}{w/o retrieved traj.} & \adjustbox{valign=c}{\firemark} Qwen3-8B
    & \textcolor{blue}{\valstd{79.0}{2.6}} & \textcolor{blue}{\valstd{49.6}{0.6}} & \textcolor{blue}{\valstd{40.7}{0.6}} \\ 
    \rowcolor{blockrow} \quad \textcolor{blue}{w/o task adaptivity} & \adjustbox{valign=c}{\firemark} Qwen3-8B
    & \textcolor{blue}{\valstd{81.2}{2.4}} &  \textcolor{blue}{\valstd{56.8}{0.4}} & \textcolor{blue}{\valstd{46.1}{0.9}} \\ 
    \rowcolor{blockrow} \quad \textcolor{blue}{w/ staged bank refresh} & \adjustbox{valign=c}{\firemark} Qwen3-8B + stage 2
    & -- & \textcolor{blue}{\valstd{61.0}{0.3}} & \textcolor{blue}{\valstd{50.5}{1.1}} \\
    \rowcolor{blockrow} \quad \textcolor{blue}{w/ test bank warm-starting} & \adjustbox{valign=c}{\firemark} Qwen3-8B
    & -- & \textcolor{blue}{\valstd{61.7}{0.4}} & \textcolor{blue}{\valstd{50.9}{0.3}} \\
    \midrule
    \multicolumn{5}{c}{\cellcolor{gray!15}\textit{Executor: GPT-5.4}} \\
    No Memory & ---
    & \valstd{62.6}{0.3} & \valstd{40.9}{0.5} & \valstd{32.6}{0.6} \\
    ReasoningBank & \adjustbox{valign=c}{\frozenmark} Qwen3-8B
    & \valstd{69.8}{4.0} & \valstd{37.4}{1.8} & \valstd{29.5}{1.0} \\
    ReasoningBank & \adjustbox{valign=c}{\frozenmark} GPT-5.4
    & \valstd{77.9}{4.2} & \valstd{43.1}{1.7} & \valstd{33.6}{1.5} \\
    MemP & \adjustbox{valign=c}{\frozenmark} GPT-5.4
    & \valstd{72.6}{1.7} & \valstd{40.8}{1.0} & \valstd{34.5}{1.7} \\
    SkillOS-base & \adjustbox{valign=c}{\frozenmark} Qwen3-8B
    & \valstd{66.9}{1.8} &  \valstd{39.7}{2.2} & \valstd{31.0}{2.0}  \\ 
    SkillOS-gpt & \adjustbox{valign=c}{\frozenmark} GPT-5.4
    & \valstd{70.0}{3.3} & \valstd{33.3}{1.1} & \valstd{26.9}{1.4} \\
    \hdashline
    \rowcolor{blockrow} \ourmethod-base  & \adjustbox{valign=c}{\frozenmark} Qwen3-8B
    & \valstd{79.3}{3.6} & \valstd{49.9}{1.0} & \valstd{39.3}{0.7} \\
    \rowcolor{blockrow} \quad \textcolor{blue}{w/o task adaptivity} & \adjustbox{valign=c}{\frozenmark} Qwen3-8B
    & \textcolor{blue}{\valstd{78.3}{6.0}} & \textcolor{blue}{\valstd{44.2}{0.5}}  &  \textcolor{blue}{\valstd{34.8}{0.8}} \\   
    \rowcolor{blockrow} \quad \textcolor{blue}{w/o successful traj.\ filtering} & \adjustbox{valign=c}{\frozenmark} Qwen3-8B
    & \textcolor{blue}{\valstd{77.6}{1.8}} & \textcolor{blue}{\valstd{45.5}{0.5}} & \textcolor{blue}{\valstd{35.9}{0.5}} \\
    \rowcolor{blockrow} \quad \textcolor{blue}{w/o raw traj.} & \adjustbox{valign=c}{\frozenmark} Qwen3-8B
    & \textcolor{blue}{\valstd{76.4}{1.5}} & \textcolor{blue}{\valstd{39.3}{0.9}} & \textcolor{blue}{\valstd{31.1}{1.0}} \\
    \rowcolor{blockrow} \ourmethod & \adjustbox{valign=c}{\firemark} Qwen3-8B
    & \valstd{86.7}{0.7} & \valstd{53.8}{0.3} & \valstd{45.4}{0.0} \\
    \rowcolor{blockrow} \quad \textcolor{blue}{w/o retrieved traj.} & \adjustbox{valign=c}{\firemark} Qwen3-8B
    & \textcolor{blue}{\valstd{72.9}{0.6}} & \textcolor{blue}{\valstd{50.0}{0.4}} & \textcolor{blue}{\valstd{40.9}{0.4}}\\
    \rowcolor{blockrow} \quad \textcolor{blue}{w/o task adaptivity} & \adjustbox{valign=c}{\firemark} Qwen3-8B
    & \textcolor{blue}{\valstd{77.4}{0.9}} & \textcolor{blue}{\valstd{48.0}{0.6}} & \textcolor{blue}{\valstd{37.5}{0.4}}  \\
    \rowcolor{blockrow} \quad \textcolor{blue}{w/ staged bank refresh} & \adjustbox{valign=c}{\firemark} Qwen3-8B + stage 2
    & -- & \textcolor{blue}{\valstd{53.6}{0.4}} & \textcolor{blue}{\valstd{46.3}{0.1}} \\
    \rowcolor{blockrow} \quad \textcolor{blue}{w/ test bank warm-starting} & \adjustbox{valign=c}{\firemark} Qwen3-8B
    & -- & \textcolor{blue}{\valstd{51.9}{0.3}} & \textcolor{blue}{\valstd{44.1}{0.4}} \\
    \bottomrule
\end{tabular}
}
\vspace{-5mm}
\end{table*}

\FloatBarrier
\textbf{Example payloads.} Below we show one curated payload per benchmark, each pairing a task with the payload the curator synthesized for it. Across all three, the curator produces task-specific guidance from retrieved trajectories. On ALFWorld, it assembles a clean-and-place procedure from three partially relevant episodes. On WebShop, it converts searches for different products into a search-phrasing and attribute-selection strategy for the current one. On $\tau^2$-bench, it distills several MMS failure cases into an ordered diagnostic procedure, separating a general resolution strategy from guidance specialized to the current request. Notably, the curator carries over the domain policy that the mutating tool requires explicit user approval.

\begin{rhbox}{ALFWorld (Task type: Clean)}
\textbf{Task}: Put a clean plate in countertop

$-----------$ Payload generated by \ourmethod $-----------$

\begin{lstlisting}
**Briefing for Task: Put a clean plate in countertop**

**Relevant Memories:**
- **Memory 1:** Successfully cooled and placed a plate on a countertop.
- **Memory 2:** Cleaned and placed a tomato on a countertop.
- **Memory 3:** Cleaned and placed a soapbar in a toilet (cleaning process is similar).

**Key Strategies:**
1. **Locate the plate:** Check countertops first, as plates are often found there.
2. **Clean the plate:** Use a sinkbasin to clean it if it's dirty.
3. **Move the plate:** Once clean, place it on a desired countertop.

**Specific Guidance:**
1. **Go to a countertop** (e.g., countertop 1 or 2) to find the plate.
2. **Take the plate** from the countertop.
3. **Go to the sinkbasin** and clean the plate if it's dirty.
4. **Return to a countertop** and place the clean plate there.
\end{lstlisting}
\end{rhbox}

\begin{rhbox}{WebShop}
\textbf{Task}: Find me high speed flashes with usb port with color: usb256-pink, and size: 512gb, and price lower than 40.00 dollars.

$-----------$ Payload generated by \ourmethod $-----------$

\begin{lstlisting}
**Briefing:**

**Target Product:** High speed flashes with usb port, color: usb256-pink, size: 512gb, price lower than $40.00

**Relevant Past Experiences:**
- **Memory 1** and **Memory 2** involved searching for men's t-shirts with specific color and size constraints, and included steps to select the correct color and size options. These experiences are relevant as they demonstrate how to navigate color and size options effectively.
- **Memory 3** involved searching for men's gym shorts with specific color and size constraints, and included steps to select the correct color and size options. This experience is also relevant as it shows how to handle color and size selection in a similar context.

**Strategies from Past Experiences:**
1. **Search Phrasing:** Use precise search terms that include the product type, key features, color, size, and price constraint. For example, "high speed flashes with usb port usb256-pink 512gb under $40".
2. **Color Selection:** Click on the exact color option from the color palette to select the desired color.
3. **Size Selection:** Click on the exact size option from the size options to select the desired size.
4. **Price Constraint:** Ensure the final product price is within the specified budget before purchasing.

**Actionable Guidance:**
1. **Search:** Begin with a search query that includes the product type, key features, color, size, and price constraint. For example, "high speed flashes with usb port usb256-pink 512gb under 40".
2. **Navigate Results:** Click on the product that matches your criteria.
3. **Select Color:** From the color options, click on "usb256-pink".
4. **Select Size:** From the size options, click on "512gb".
5. **Verify Price:** Ensure the final price is within the specified budget before proceeding to purchase.
\end{lstlisting}
\end{rhbox}

\begin{rhbox}{$\tau^2$-bench (Domain: Telecom)}
\textbf{Task}: You are unable to send MMS messages using your messaging app for the past few hours. You want to fix it and successfully send an MMS message.

$-----------$ Payload generated by \ourmethod $-----------$

\begin{lstlisting}

Most relevant memories: **Memory 1 and Memory 3** (same MMS issue, including travel/roaming and possible data-cap blockage). **Memory 2** is also useful for the domestic case where app permissions/network mode were the fix.

Resolution strategy that worked:
1. **Identify the customer/line**
   - Use `get_customer_by_phone` (or equivalent customer lookup).
   - If multiple lines exist, confirm the exact phone number/line affected.

2. **Gather device/network state first**
   - Ask user to run/read: `check_network_status`, `can_send_mms`, `check_apn_settings`, `check_wifi_calling_status`, and app permission checks (use `check_installed_apps` first if app name is unclear).
   - If abroad, specifically verify **mobile data ON**, **data roaming ON**, and usable network service (not airplane mode / no SIM / 2G-only).

3. **Guide user through non-account fixes in order**
   - If airplane mode is on: turn it off.
   - If SIM missing: reseat SIM.
   - If mobile data or roaming is off: turn them on.
   - If stuck on poor/2G service: set preferred network mode to `4g_5g_preferred`.
   - If messaging app lacks permissions: grant **sms** and **storage**.
   - If needed, check/reset APN and reboot.
   - If Wi-Fi Calling is on and MMS still fails, try turning it off.

4. **Check account-side blockers only after device state is good**
   - Look up the correct line with `get_details_by_id`.
   - Verify line is active and whether roaming is enabled on the account.
   - Use `get_data_usage` and plan lookup to see if the line is over its data limit.

5. **If over data limit and MMS depends on mobile data**
   - Explain the overage and available options.
   - Confirm allowed refuel amount and total charge.
   - Obtain explicit user approval.
   - Apply with `refuel_data`.

Guidance for this request:
- Start by identifying the customer/line, then immediately collect current device state with:
  - `check_network_status`
  - `can_send_mms`
  - `check_apn_settings`
  - `check_wifi_calling_status`
  - `check_installed_apps` -> `check_app_permissions` for the actual messaging app
- Based on results, prioritize:
  1. restore basic cellular service (airplane mode/SIM/signal/network mode),
  2. ensure mobile data is enabled,
  3. if roaming is relevant, ensure device roaming is on and account roaming is enabled,
  4. fix app permissions (`sms`, `storage`),
  5. APN reset/reboot if still failing.
- If the device shows good service/data and MMS still fails, check the account line and **data usage**; if at/over limit, offer a **data refuel** only after confirming price and receiving explicit approval before calling the mutating tool.
- Do **not** assume the same line ID, roaming status, or data usage from memory-look up the current case.
\end{lstlisting}

\end{rhbox}

\begin{figure*}[!th]
\centering
    \includegraphics[width=0.7\linewidth]{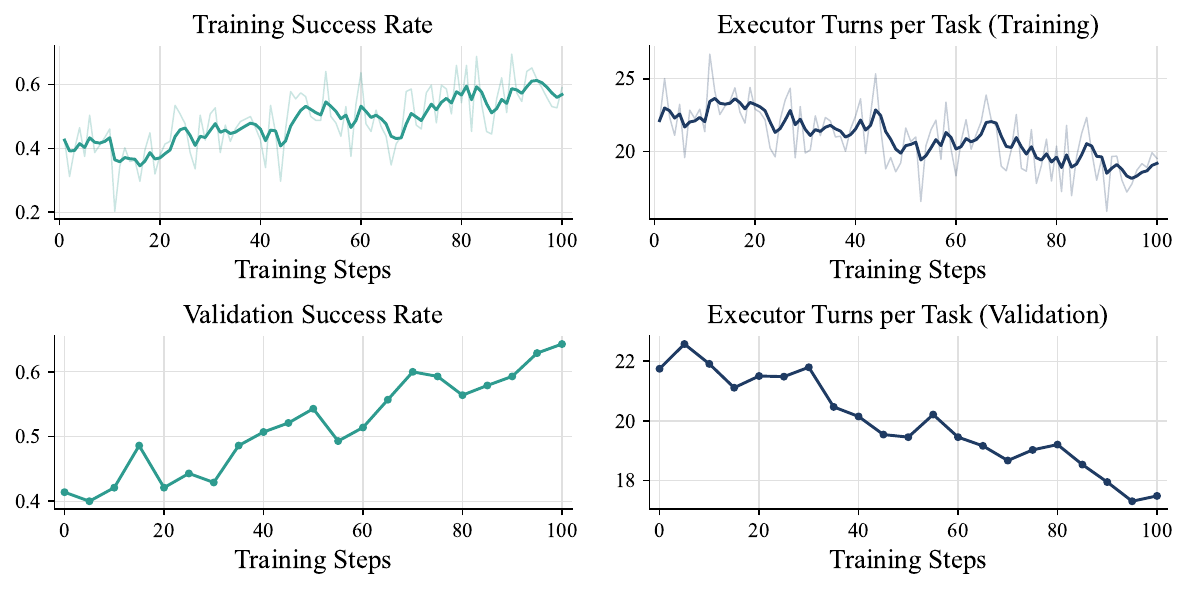}
    \caption{GRPO training curves on ALFWorld (Qwen3-8B executor). Top row: training SR and executor turns. Bottom row: the same metrics on validation. SR rises and turns fall steadily over 100 steps.\looseness-1}
    \vspace{-4mm}
    \label{fig:training-curve-alfworld}
\end{figure*}

\begin{figure*}[!th]
\centering
    \includegraphics[width=\linewidth]{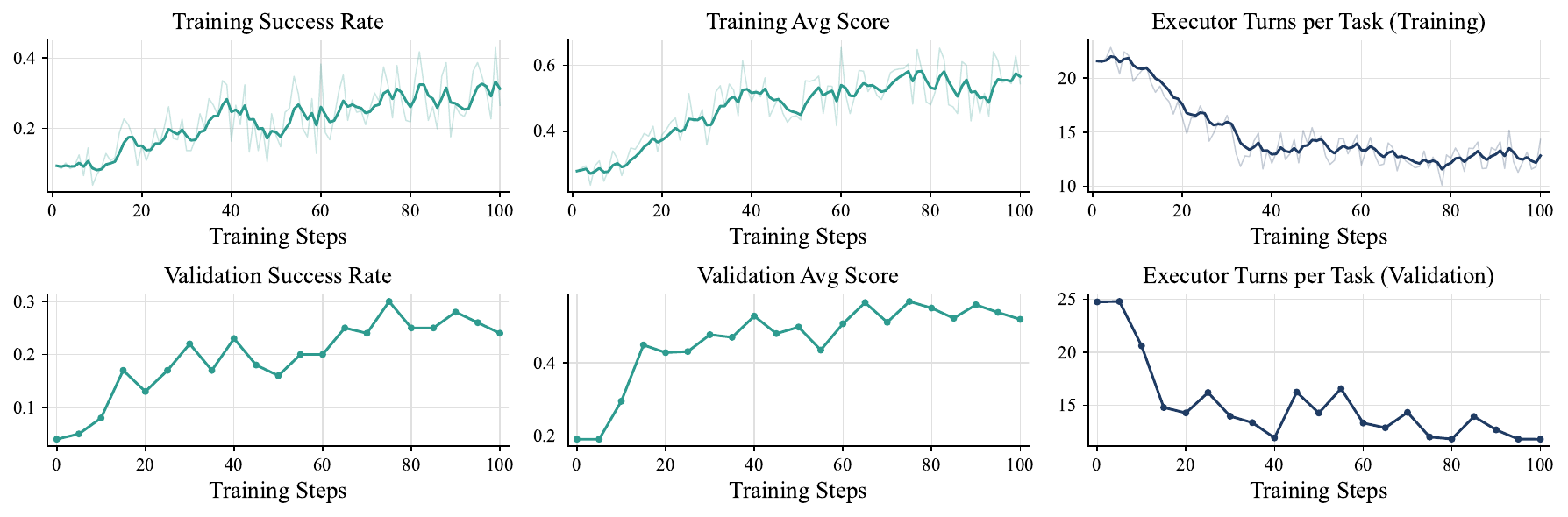}
    \caption{GRPO training curves on WebShop (Qwen3-8B executor). Top row: training SR, training score, training executor turns. Bottom row: the same three metrics on validation.}
    \label{fig:training-curve-webshop}
\end{figure*}

\textbf{Training curves.} \Cref{fig:training-curve-alfworld,fig:training-curve-webshop} show GRPO training progress on ALFWorld and WebShop, with validation on a subsampled test stream. On both benchmarks, validation success rate climbs steadily over 100 training steps while executor turns per task fall, indicating the curator learns to produce payloads that improve task success while reducing the number of executor steps. Training is stable throughout under a single task reward, with no auxiliary content-quality reward, no task grouping, and no return shaping — supporting the claim that read-time curation makes curator learning straightforward.

\end{document}